\documentclass{article}
\usepackage{comment}

\usepackage{subcaption}
\usepackage{amssymb}
\usepackage{stfloats}
\usepackage{ijcai24}
\usepackage{times}
\usepackage{soul}
\usepackage{url}
\usepackage[hidelinks]{hyperref}
\usepackage[utf8]{inputenc}
\usepackage{graphicx}
\usepackage{amsmath}
\usepackage{amsthm}
\usepackage{booktabs}
\usepackage{algorithm}
\usepackage{algorithmic}
\usepackage[switch]{lineno}

\usepackage{xcolor}

\nolinenumbers

\title{Progressive trajectory matching for medical dataset distillation}

\author{
Zhen Yu$^1$
\and
Yang Liu$^2$\and
Qingchao Chen$^{1*}$\\
\affiliations
$^1$National Institute of Health Data Science, Peking University\\
$^2$Wangxuan Institute of Computer Technology, Peking University\\
}

\begin{document}

\maketitle

 \begin{abstract}
It is essential but challenging to share medical image datasets due to privacy issues, which prohibit building foundation models and knowledge transfer. In this paper, we propose a novel dataset distillation method to condense the original medical image datasets into a synthetic one that preserves useful information for building an analysis model without accessing the original datasets. Existing methods tackle only natural images by randomly matching parts of the training trajectories of the model parameters trained by the whole real datasets. However, through extensive experiments on medical image datasets, the training process is extremely unstable and achieves inferior distillation results. To solve these barriers, we propose to design a novel \textit{progressive trajectory matching strategy} to improve the training stability for medical image dataset distillation. Additionally, it is observed that improved stability prevents the synthetic dataset diversity and final performance improvements. Therefore, we propose a dynamic overlap mitigation module that improves the synthetic dataset diversity by dynamically eliminating the overlap across different images and retraining parts of the synthetic images for better convergence. Finally, we propose a new medical image dataset distillation benchmark of various modalities and configurations to promote fair evaluations. It is validated that our proposed method achieves $8.33\%$ improvement over previous state-of-the-art methods on average, and $11.7\%$ improvement when $ipc=2$ (\textit{i.e.,}, image per class is $2$).
Codes and benchmarks will be released.
    
\end{abstract}
 \section{Introduction}


\begin{figure}[htbp]
        \centering
        \subcaptionbox{Overview\label{a}}{
        \includegraphics[width=1.645in]{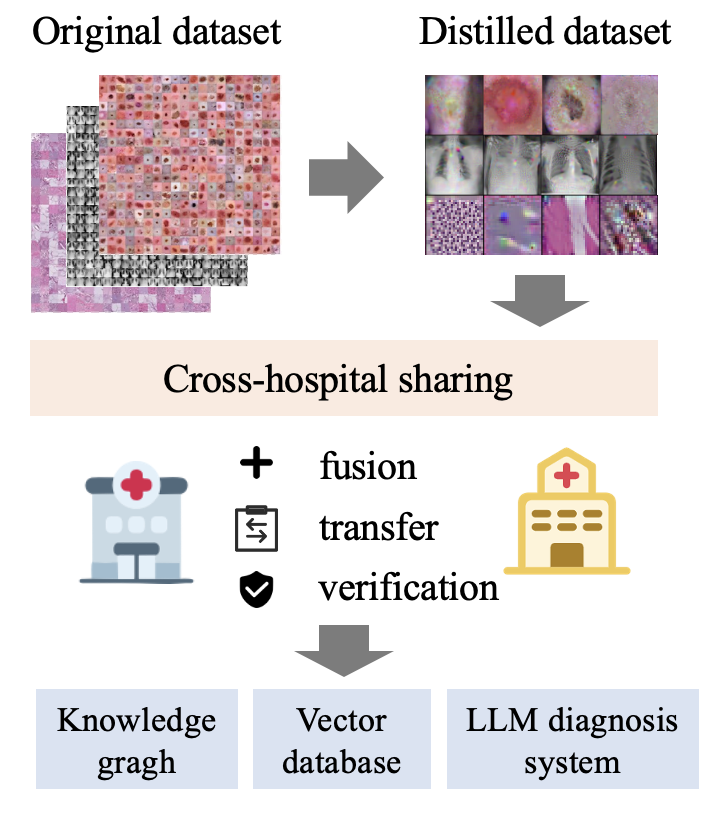}
    }\hspace{-0.15cm}
        \subcaptionbox{Performance\label{b}}{
        \centering
        \includegraphics[width=1.6in]{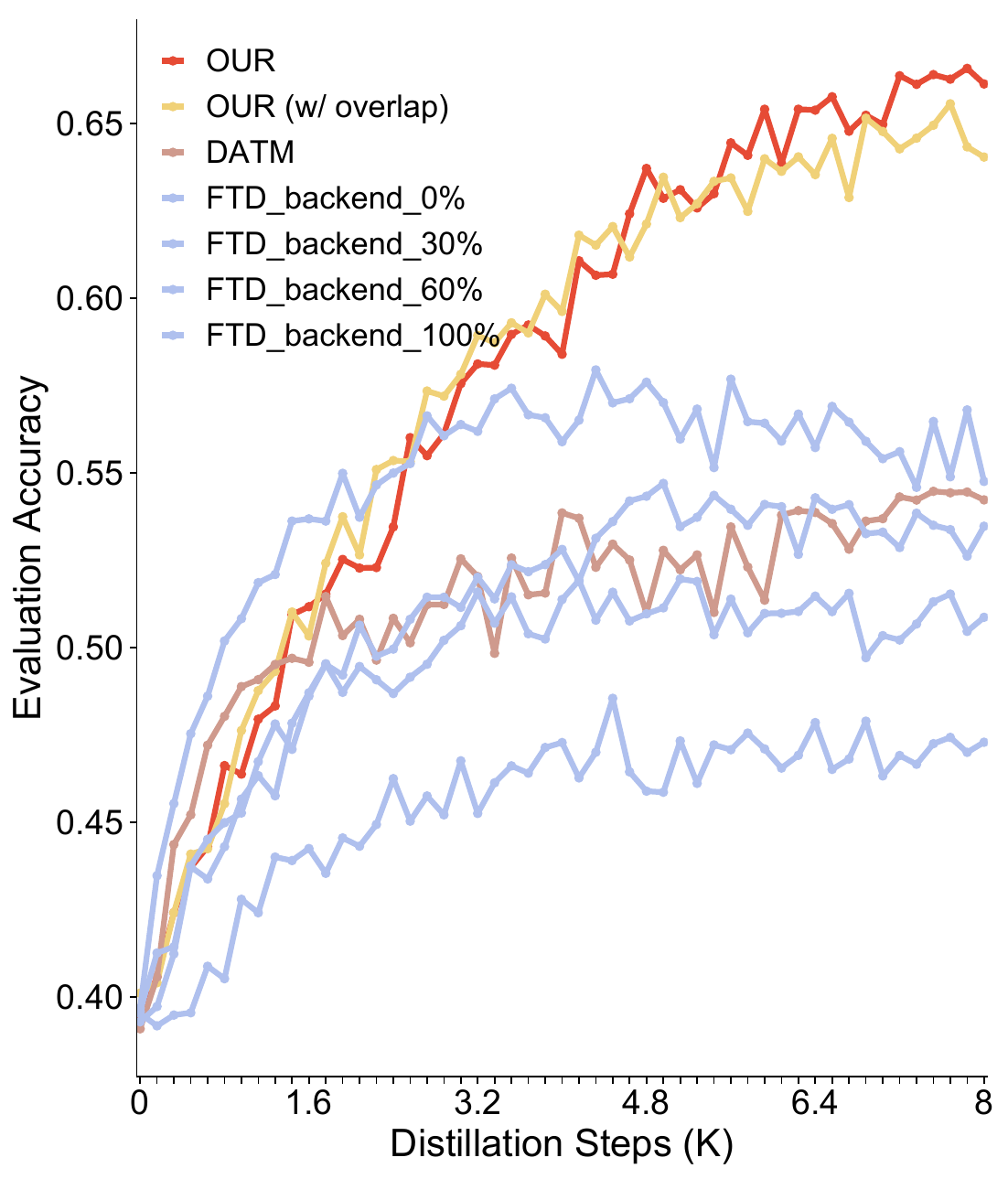}
    }
        \caption{(a) overview for medical dataset distillation.
        (b) performance plot of various methods and using different probabilities of sampling the backend trajectories. }
        \label{fig:000}
\end{figure}

Effective accumulation, sharing, and management of large-scale medical datasets across multi-cohorts/hospitals is key to improving the quality of medical services and research. However, issues such as privacy protection, data storage, transmission, preprocessing, and computational costs are extremely challenging. Dataset distillation~\cite{geng2023survey,yu2023dataset,sachdeva2023data,lei2023comprehensive} seems a good candidate to solve previous barriers, which aims to condense a given dataset into a synthetic one, that preserves the information of the original one. Meanwhile training on the synthetic ones is expected to achieve similar performance compared with training on the original dataset. 




Although existing dataset distillation methods have been evaluated on natural image datasets, including distribution-based matching in feature space and trajectory-based matching in parameter space, their effectiveness has not been comprehensively understood/verified for distilling medical image datasets. To our best knowledge, only \cite{li2022dataset} follows the distillation paradigm of MTT~\cite{cazenavette2022dataset} and verified its effectiveness on sorely the chest X-ray images. Compared with natural images, as both the medical images and the analysis task exhibit extremely different characteristics and technical challenges, it is of the essence to establish a  comprehensive benchmark for the medical image dataset distillation and identify the challenges with potential solutions. 

We proposed a new benchmark, MDD with stats in Table \ref{table_1}, composed of six public medical datasets, covering a variety of modalities, analysis tasks, and resolutions. Through comprehensive analysis of the state-of-the-art methods as in Table \ref{table_6} and Table \ref{table_2}, we articulate \textit{the summarized challenges in medical dataset distillation as follows.}


\textbf{Degraded training stability of existing matching trajectory segments randomly.} 
As shown in Figure \ref{b}, we observed in medical image distillation that: (1) it is essential to match and distill the beginning/front-end part of expert trajectories; (2) random matching of trajectory segments fails to achieve stabilized and results of superior performances. As shown by the bottom four lines in Figure~\ref{fig:000}, they represent the probabilities of randomly matching the back-end trajectories at $0\%$, $30\%$, $60\%$ and $100\%$ respectively, and it can be observed that randomly matching more back-end trajectories results in poorer performance. In addition, even if the probability is $0$, the lack of back-end trajectory information results in a significant performance gap compared to our method from the up two lines in Figure~\ref{fig:000}.

\textbf{Lack of the diversity of synthetic images and saturated gradients in optimizations.} As shown in Figure 1, when there are multiple synthetic images per class in the budget, the existing distillation methods tend to obtain similar synthetic images, a.k.a, their overlaps are high. This overlap is caused by the gradient update of the synthetic images as a whole, resulting in more saturated gradients of similar synthetic images as the distillation progresses. Lack of diversity in the synthetic ones tends to achieve inferior performances. We also observe using experiments that the overlap restricts the performance improvement at the end of distillation (in supplementary).



To address the aforementioned two challenges, we first design a novel progressive trajectory matching strategy to promote the stability of multi-step trajectory matching. Promoting stability incurs the lack of synthetic image diversity. To solve this, we propose a novel overlap prevention module, that maintains the diversity of synthetic images within classes and improves the convergence using the retraining strategy.



Our contributions are summarized as follows:
\begin{itemize}
    \item We propose a new and comprehensive benchmark to evaluate the medical image dataset distillation. This benchmark covers a variety of conditions, including imaging modalities, image resolutions, and analysis tasks. Through this benchmark, we identify the challenges to distilling the medical image datasets.
    \item We propose a novel progressive trajectory matching strategy, scheduling the start and end points of the trajectory. This strategy proves to significantly improve the benchmark results.
    \item Stability incurs the diversity of learned synthetic datasets. We propose a novel module to prevent synthetic image overlaps and design a retraining strategy for better convergence. 
    \item Our proposed method achieves the state-of-the-art results evaluated in various benchmarks.
\end{itemize}




 \section{Related Works}

To our best knowledge, dataset distillation was first formally proposed by ~\cite{wang2018dataset} to synthesize a smaller synthetic dataset, so that once trained, it can match the test accuracy using the complete dataset. The mainstream framework can be further divided into distribution-based and trajectory-based matching methods.

\textbf{Distribution Matching Methods:}
Distribution matching learns a synthetic dataset by approximating its feature distribution to the real feature distribution. The commonly used distance function includes metrics such as maximum mean difference (MMD). DM~\cite{zhao2023dataset} uses a classification encoder to extract features and pulls features of each class closer between the synthetic and real datasets. CAFE~\cite{wang2022cafe} further improves the DM by aligning the layer-wise features formalized as a dynamic bi-level optimization. It seems to better capture the distribution of the entire dataset. Although DM ignores optimization steps in the parameter space matching and alleviates the memory requirements, imprecise estimation of the high-dimensional data distribution often leads to inferior performance. \textit{Different from them, our work designs a novel strategy to progressively match the trajectory of parameter space.}


\textbf{Single-step Trajectory Matching:}
The key idea under this regime is to train the synthetic parameters and synthetic data simultaneously by improving the consistency between the synthetic parameters and the buffer one trained by real datasets. This regime can be further divided into single-step and multi-step trajectory matching. DC~\cite{zhao2020dataset} may be the first single-step matching method by maintaining consistent single-step gradients. Subsequently, DSA~\cite{zhao2021dataset} improves the performance of DC by incorporating image augmentation functions in training(\textit{e.g.,}scale, flip, etc).
As these image enhancement methods are universally applicable, DSA has become a common part of all subsequent methods. \textit{Different from them, our work belongs to the multi-step trajectory matching.}

\textbf{Multi-step Trajectory Matching:}
MTT~\cite{cazenavette2022dataset} may be the first multi-step matching method, which matches the long-range training dynamics.
SMDD~\cite{li2022dataset} follows the MTT and prunes those difficult-to-match parameters, ensuring the robustness of the synthesized dataset.
FTD~\cite{du2023minimizing} proposed to regularize the flatness of multiple trajectories, to avoid the accumulation of trajectory-matching errors and improve the network robustness. DATM~\cite{guo2023towards} finds that the training stage of the selected matching trajectory (\textit{i.e.,} early or late stages) greatly affects the performance, and proposes using early trajectory for low-image per class (IPC) 
and later trajectory for high-IPC. SeqMatch~\cite{du2023sequential} finds it difficult to learn high-level features for high-IPC scenarios and proposes to divide the synthetic dataset into multiple subsets
and distill each subset from the different stages of trajectories. From the perspective of \textit{medical dataset distillation}: existing medical dataset distillation methods~\cite{li2020soft,li2022compressed,li2022dataset} are only verified on the X-ray modality and follow MTT. 
\textit{Our work designs a novel strategy to progressively match the trajectory rather than randomly as the previous do. In addition, we proposed a novel overlapping prevention module involving the re-training strategy, while they only implicitly pre-set different subsets to update as SeqMatch did. Our methods exhibit less computational complexities. Besides all the method differences, we established a new medical dataset distillation benchmark enabling more diverse and comprehensive evaluations.}



 \section{Methods}

\subsection{Overall Framework}

We define the original dataset and synthetic dataset as $\mathcal{D}=\left\{\left(x_i, y_i\right)\right\}_{i=1}^{|\mathcal{D}|}$ and $\mathcal{S}=\left\{\left(s_i, y_i\right)\right\}_{i=1}^{|\mathcal{S}|}$ ($|\mathcal{S}| \ll |\mathcal{D}|$) respectively, where the data samples $x_{i}, s_{i} \in \mathbb{R}^d$, the class labels $y_i \in \mathcal{Y}=\{0,1, \ldots, C-1\}$ and $C$ is the number of classes. 
Each class of $\mathcal{S}$ contains ipc (images per class) data pairs, and we evenly divide $\mathcal{S}$ into foundation subset $\mathcal{S}^{f}$ and complement subset $\mathcal{S}^{c}$. 
\textit{i.e.,} $|\mathcal{S}|= |\mathcal{S}^{f}|+|\mathcal{S}^{c}| = \mathrm{ipc} \times C $ and $|\mathcal{S}^{f}| = |\mathcal{S}^{c}|$.
we define expert parameters $\theta^{\mathcal{D}}_t$
as teacher network parameters trained on full images at the training step $t$, and synthetic parameters $\theta_t^{\mathcal{S}}$ as the student network parameter trained on synthetic images,
and a complete expert trajectory and synthetic trajectory can be represented as $\left\{\theta^{\mathcal{D}}_t\right\}_{t=0}^T$ and $\left\{\theta^{\mathcal{S}}_t\right\}_{t=0}^T$, where $T$ denotes the training steps.

The overall architecture is shown in Figure~\ref{overall}. 
The goal of dataset distillation is to learn a synthetic dataset that \textit{mimics the model performances using a real dataset} if training under the same configurations. To achieve this, we adopt the bi-level optimization framework of multi-step trajectory matching, where we simultaneously learn the synthetic student network trajectories using cross-entropy loss $\mathcal{L}_c$, meanwhile updating the synthetic images by matching the parameter trajectories between the synthetic student network and the buffer teacher network using $\mathcal{L}_{match}$.

As demonstrated in both analysis and introductions, existing trajectory matching methods experience \textit{degraded training stability} on medical image datasets. To improve it, as shown in Figure 2, we propose a novel progressive matching strategy by scheduling the start point and end point of each synthetic trajectory under match. Specifically, instead of the random start/end point as existing methods did, we propose to \textit{set the very beginning of the network parameters as the start point in each distillation step}. Additionally, we propose to \textit{progressively increase the endpoint of the trajectory and reuse the intermediate synthetic images as in each start point}
, which allows for the gentle learning of hard information in the back-end trajectory without disrupting the crucial early-stage information.
More details are in section 3.2. 

Improved stability strategies tend to promote similarities/overlaps among the distilled images of the same class. Therefore, the lack of diversity becomes an issue for improved performances and saturated gradients at the end of the distillation procedure (see supplementary). To solve it, as shown in Figure 2, we propose a novel image overlap prevention module, with a re-training strategy to improve the convergence using $\mathcal{L}_{overlap}$. More details are in section 3.3.

\subsection{Progressive Trajectory Matching}
The reasons for the inferior performance of random trajectory matching methods include: (1) the front-end information from the multi-expert trajectory is more stable and reliable compared to the back-end, which fails to utilize multi-expert.
(2) the fitting of small segments between trajectories is 
fragmented and rigid, and classification loss can seriously conflict with matching loss. (3) learning from back-end trajectories can improve the recognition of hard samples, but it is important to ensure that previous information is not forgotten or overwritten.

The progressive matching mechanism we designed has the following characteristics: (1) Each distillation starts from $0$, \textit{i.e.,} $\theta_0^\mathcal{S}: = \theta_0^\mathcal{D} $, progressively and non-forgetfully conducting ordered learning from easy to difficult.
(2) The step size $T$ for each distillation gradually increases from $1$ to $T$, which can greatly alleviate the conflict between classification and matching.
(3) Our method can achieve a certain distillation effect in a very small time and storage cost, and even if trained to the last $T$ step, our cost is also comparable to random matching.
therefore, $\mathcal{S}$ can be an approximate solution of the following optimization problem:
\begin{align}
\label{math:2}
\mathcal{S} = \underset{\mathcal{S} \subset \mathbb{R}^d \times \mathcal{Y},|\mathcal{S}|=\mathrm{ipc} \times C}{\arg \min } \underset{\theta_0 \sim P_{\theta_0}}{\mathbb{E}}\left[\sum_{t=0}^{T} \frac{\mathbf{D}\left(\theta_{t}^{\mathcal{D}}, \theta_{t}^{\mathcal{S}}\right)}{\mathbf{D}\left(\theta_{0}^{\mathcal{D}}, \theta_t^{\mathcal{D}}\right)}\right]
\end{align}
where $\mathbf{D}$ is a distance metric of choice and $t$ is a gradually increasing trajectory step size. Each round of progressive matching reduces the distance between synthetic trajectory and multi-expert trajectories, and this fitting process is stable and orderly.

\begin{figure}[t]
 \centering
  \includegraphics[width=4.0in]{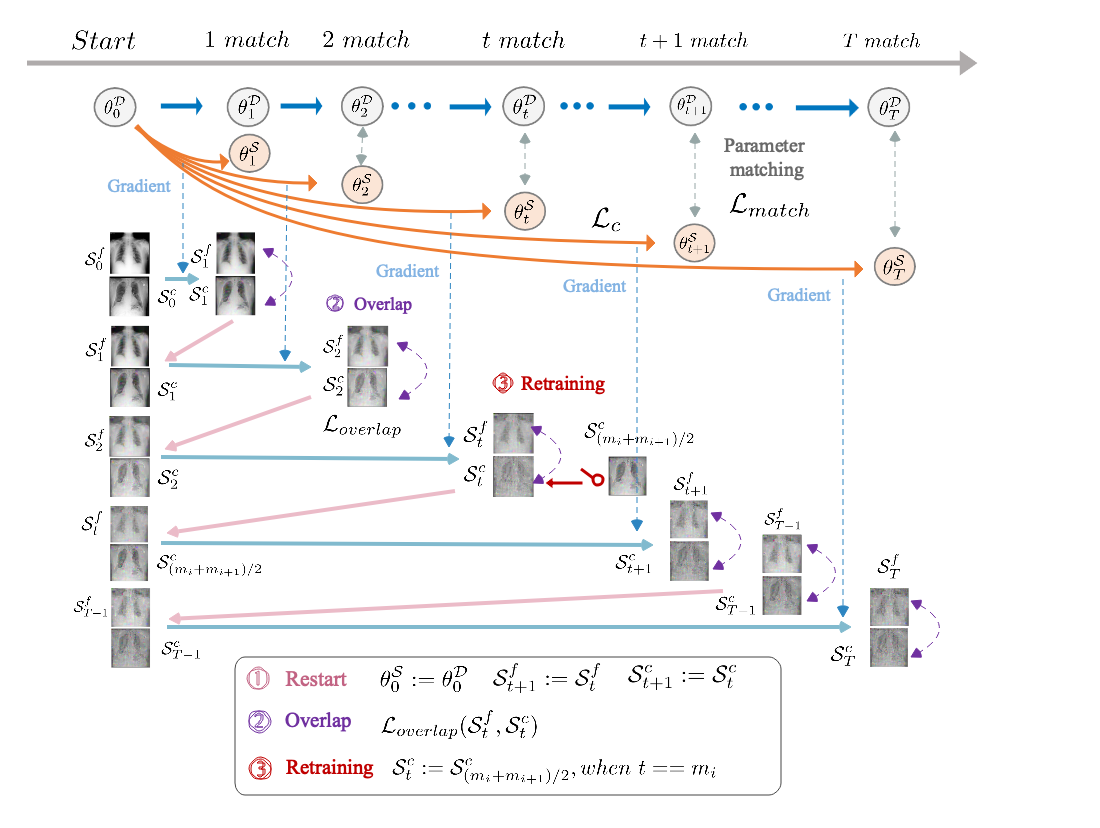}
 \caption{ Overall architectures. Using multiple buffer trajectories, the synthetic images and the labels are used to minimize cross-entropy loss and obtain the orange synthetic student trajectory parameters. By comparing the differences between synthetic and buffer trajectories, the gradients are back-propagated to the synthetic images and they are updated. We propose progressive trajectory matching, by iteratively going back to the original starting points for stable matching. We propose dynamic overlap mitigation and retraining techniques to improve synthetic image diversity.   
}
 \label{overall}
\end{figure}

As shown in Figure~\ref{overall}, at the beginning of each distillation, we first sample initial parameters $\theta_0^\mathcal{D}$ from multi-expert trajectories and use $\theta_0^\mathcal{D}$ to initialize student parameters $\theta_0^\mathcal{S}$. 
Then, gradient descent updates are executed on the student parameters by reducing the classification loss of the synthetic data:
\begin{align} 
\theta_{t}^{\mathcal{S}} = \theta_{0}^{\mathcal{S}}-\eta \cdot  
\nabla \ell\left(\mathcal{A}\left(\mathcal{S}\right) ; \theta_{0}^{\mathcal{S}}\right)
\end{align}
where $\mathcal{A}$ is a differentiable data augmentation module proposed in DSA~\cite{zhao2021dataset}, and $\eta$ is the learning rate used to update the student network.
Finally, at the end of each distillation, we update synthetic images $\mathcal{S}$ based on the parameter matching loss between $\theta_{t}^\mathcal{S}$ and $\theta_{t}^\mathcal{C}$:
\begin{align} 
\mathcal{L}_{match}=\frac{\left\|\theta^{\mathcal{S}}_{t}-\theta_{t}^{\mathcal{D}}\right\|_2^2}{\left\|\theta_t^{\mathcal{D}}-\theta_{0}^{\mathcal{D}}\right\|_2^2}
\end{align} 
the updated synthetic images return to the initial stage and randomly sample expert trajectory parameters while increasing the step size for the next round of trajectory matching.

\subsection{Dynamic Retraining and Overlap Mitigation}

The progressive trajectory matching extracts more sufficient preliminary information from multi-experts, which is crucial for medical datasets. However, it introduces the issue of overlap, leading to lack of diversity.
To address this, 
we divide $\mathcal{S}$ into foundation subset $\mathcal{S}^{f}$ and complement subset $\mathcal{S}^{c}$, and
amplify the distribution differences between $\mathcal{S}^{f}$ and $\mathcal{S}^{c}$ to enhance the diversity of synthetic images. 

We proposed to use the maximum mean discrepancy (MMD)~\cite{gretton2012kernel} to measure the distribution distance between the foundation subset $\mathcal{S}^{f}$ and complement subset $\mathcal{S}^{c}$. 
We define the overlap loss to minimize as:
\begin{align} 
\mathcal{L}_{overlap}= \sum_{i=1}^{|\mathcal{S}^{f}|} \sum_{j=1}^{|\mathcal{S}^{c}|} \left(2 k\left({\mathcal{S}^{f}_i}, {\mathcal{S}^{c}_j}\right) -k\left({\mathcal{S}^{c}_i} , {\mathcal{S}^{c}_j}\right) -
k\left({\mathcal{S}^{f}_i}, {\mathcal{S}^{f}_j}\right)
\right)
\end{align}
where $k(a, b)=\exp \left(-\frac{\|a-b\|^2}{2 \sigma^2}\right) \cdot \mathbb{I}(a, b)$ denotes the kernel function, and $\mathbb{I}(a, b)$ is an indicator function, taking the value $1$ when the two items under comparison come from the same class and $0$ otherwise.

Although overlap mitigation increases the diversity of synthetic images, we empirically found that its loss increases exponentially and is difficult to converge, as shown by the blue line in Figure \ref{fig:3:c}. To tackle this, we designed a retraining mechanism by replacing the complementary synthetic images with their historical ones and continuing the training as shown in Figure 2. 
Specifically, we divide the whole distillation procedure ranging from $0$ to $T$ proportionally into $k$ retraining points, which are labeled as $m_i$ and the set is defined as $M=\left\{m_1, m_2, \ldots, m_k\right\}$, and $m_i$ is denoted as: $m_i=T-\frac{T}{k+1} \cdot i$. 
As shown in Figure~\ref{overall},
when the distillation reaches the $t^{th}$ match, we verify whether $t$ is within the retraining time point set $M$, \textit{i.e.,} $\mathbb{I}(\exists \ i: m_i = t) $ is equal to $1$.
We only re-train the complement subset $\mathcal{S}^{c}$ when it reaches the retraining point and replaces the synthetic image as the one in the middle of the previous retraining interval.
\textit{i.e.,} $\mathcal{S}^{c}_t: = \mathcal{S}^{c}_{(m_i+m_{i+1})/2}$. 

\subsection{Overall Optimization}
To sum up, the overall optimization is as follows, where $\beta_1$ and $\beta_2$ are hyperparameters.

\begin{align}
\min_{\mathcal{S},\theta^{\mathcal{S}}} \mathcal{L}_{c} + \beta_1 \mathcal{L}_{match}  + \beta_2 \mathcal{L}_{overlap}.
\end{align}

\begin{table*}
\footnotesize
\centering
\caption{Statistics of medical datasets used for the proposed benchmark, with comprehensive coverage.}
\renewcommand{\arraystretch}{1.08}
\label{table_1}
\renewcommand{\arraystretch}{1.15} 
\begin{tabular}{c|c|c|c|c|c}
\hline
\textbf{Dataset Name} & \textbf{Data Modality}  & \textbf{2D/3D}  & \textbf{Classes Number} & \textbf{Sample Number} & \textbf{Image Shape} \\
\hline
PATHMNIST~\cite{kather2019predicting} & Colon Pathology & 2D & Multi-Class (9)  & 107,180  & $3\times28\times28$ \\
OCTMNIST~\cite{kermany2018identifying} & Retinal OCT & 2D & Multi-Class (4)  & 109,309  & $1\times28\times28$ \\
ORGAN3D~\cite{bilic2023liver} & Abdominal CT & 3D & Multi-Class (11)  & 1,743  & $1\times28\times28\times28$ \\
COVID19-CXR~\cite{rahman2021exploring} & Chest X-ray & 2D & Multi-Class (4)  & 21,165  & $3\times300\times300$ \\
SKIN-HAM~\cite{tschandl2018ham10000} & Dermatoscope & 2D & Multi-Class (7)  & 10,015  & $3\times600\times450$ \\
BREAST-ULS~\cite{al2020dataset} & Ultrasound & 2D & Multi-Class (3)  & 780  & $3\times500\times500$ \\
\hline
\end{tabular}
\end{table*}
\begin{table*}
\caption{Performances of ConvNet using various distillation methods on the \textbf{Mnist 2D/3D} medical datasets.}
\label{table_6}
\renewcommand{\arraystretch}{1.2} 
\resizebox{0.95\linewidth}{!}{
\begin{tabular}{c|cc|cc|cc}
\hline & \multicolumn{2}{|c}{ PATHMNIST $(28\times28)$ } & \multicolumn{2}{c}{ OCTMNIST $(28\times28)$} & \multicolumn{2}{c}{ ORGAN3D $( 28\times28\times28)$ } \\
IPC & 2  & 10 & 2 & 10 & 2 & 10 \\  
\hline Real Dataset & \multicolumn{2}{|c}{$88.48 \pm 0.02$} &  \multicolumn{2}{c}{$65.60 \pm 0.02$} & \multicolumn{2}{c}{$81.31 \pm 0.04$} \\  
\hline 
DC~\cite{zhao2020dataset} & $57.63 \pm 0.01$ & $69.79 \pm 0.01$ & $42.15 \pm 0.02$ & $60.00 \pm 0.01$ & $49.87 \pm 0.02$ & $74.13 \pm 0.01$\\
DM~\cite{zhao2023dataset} & $50.74 \pm 0.02$ & $78.20 \pm 0.01$ & $38.42 \pm 0.02$ & $58.84 \pm 0.01$ & $49.03 \pm 0.02$ & $75.07 \pm 0.01$ \\
DSA~\cite{zhao2021dataset}  & $58.56 \pm 0.02$ & $77.07 \pm 0.01$ & $48.56 \pm 0.02$ & $60.91 \pm 0.01$ & $50.13 \pm 0.02$ & $74.82 \pm 0.01$ \\
CAFE~\cite{wang2022cafe} & $62.03 \pm 0.02$ & $73.17 \pm 0.01$ & $41.84 \pm 0.01$ & $62.67 \pm 0.01$ & $51.15 \pm 0.02$ & $76.07 \pm 0.01$ \\
MTT~\cite{cazenavette2022dataset} & $50.86 \pm 0.02$ & $69.47 \pm 0.01$ & $36.10 \pm 0.01$ & $54.67 \pm 0.01$ & $52.20 \pm 0.01$ & $69.56 \pm 0.01$ \\
SMDD~\cite{li2022dataset}  & $51.53 \pm 0.02$ & $70.11 \pm 0.01$ & $37.63 \pm 0.01$ & $53.82 \pm 0.01$ & $52.46 \pm 0.02$ & $70.27 \pm 0.01$ \\
HaBa~\cite{liu2022dataset} & $49.10 \pm 0.04$ & $71.54 \pm 0.01$ & $35.18 \pm 0.03$ & $58.23 \pm 0.01$ & $51.54 \pm 0.01$ & $70.49 \pm 0.01$ \\
FTD~\cite{du2023minimizing} & $53.47 \pm 0.02$ & $70.28 \pm 0.01$ & $42.61 \pm 0.02$ &  $57.04 \pm 0.01$ & $53.13 \pm 0.01$ & $71.34 \pm 0.01$ \\
DATM~\cite{guo2023towards}  & $54.75 \pm 0.02$ & $72.81 \pm 0.01$ & $43.67 \pm 0.02$ & $60.48 \pm 0.01$ & $54.07 \pm 0.01$ & $72.42 \pm 0.01$ \\
SeqMatch~\cite{du2023sequential} &$46.30 \pm 0.01$ & $72.54 \pm 0.01$ & $35.70 \pm 0.01$ & $59.21 \pm 0.01$ & $48.92 \pm 0.01$  & $71.89 \pm 0.01$ \\
\hline 
OUR & $\mathbf{67.01} \pm \mathbf{0.01}$ & $\mathbf{74.42} \pm \mathbf{0.01}$ & $\mathbf{49.59} \pm \mathbf{0.03}$ & $\mathbf{61.84} \pm \mathbf{0.01}$ & $\mathbf{56.28} \pm \mathbf{0.01}$ & $\mathbf{73.95} \pm \mathbf{0.01}$ \\
\hline
\end{tabular}
}
\end{table*}


 \section{Experiments}


\subsection{Benchmark and Datasets}

Due to the data modalities in medical datasets are diverse, the data dimensions are uncertain and the resolution is high, making it difficult to ensure that successful distillation methods in natural images can be well adapted to medical images. Therefore, a comprehensive medical dataset benchmark is needed to evaluate the performance of dataset distillation methods.
We retrieved and divided public medical datasets 
~\cite{yang2023medmnist,kather2019predicting,kermany2018identifying,bilic2023liver,rahman2021exploring,tschandl2018ham10000,al2020dataset} 
into different and multiple subsets, aiming to clarify the evaluation objectives of each subset and cover a wide range of medical application scenarios. 
We only evaluate classification tasks and more complex tasks such as image detection and segmentation remain untapped, and this is a step-by-step process that we will explore next.
The classification performance of all public datasets has been validated as effective in~\cite{yang2023medmnist},
and the detailed information is shown in Table~\ref{table_1}.

\subsubsection{High-Resolution Medical Benchmark}

\textbf{COVID19-CXR}~\cite{rahman2021exploring} is a chest X-ray dataset for COVID-19, consisting of 3616 COVID-19 positive cases along with 10192 normal, 6012 Lung Opacity and 1345 viral pneumonia images, which can serve as an excellent benchmark for high-density structure sensitive medical images.
\textbf{SKIN-HAM}~\cite{tschandl2018ham10000} is a large collection of multi-source dermatoscopic images of pigmented lesions from different populations, which consists of 10,015 images with imbalanced distribution and can be considered as a suitable benchmark for dealing with imbalanced medical dataset.
\textbf{BREAST-ULS}~\cite{al2020dataset} is a breast ultrasound dataset for breast cancer, which consists of 780 breast ultrasound images and is categorized into normal, benign and malignant, and can be used as a medical benchmark for high-resolution and low sample data scale.

\subsubsection{Mnist 2D/3D Medical Benchmark}

\textbf{PATHMNIST}~\cite{kather2019predicting} is a 9-class colon pathology dataset for predicting survival from colorectal cancer histology slides, which is resized to $3\times28\times28$ similar to natural image datasets (CIFAR, Tiny ImageNet, \textit{etc.}) and can serve as the standard MNIST dataset benchmark. 
\textbf{OCTMNIST}~\cite{kermany2018identifying} is a 4-class valid optical coherence tomography images for retinal diseases, which can be used as a single channel grayscale image benchmark in medical data. 
\textbf{ORGAN3D}~\cite{bilic2023liver} is a 11-class 3D computed tomography (CT) images from liver tumor segmentation dataset, and 3D bounding boxes are used to process these CT images into $1\times28\times28\times28$ for body organ classification, which can be regarded as an appropriate benchmark for unique 3D data (CT, MRI, \textit{etc.}) in medical field.

\subsection{Baselines and Models}
We compare our method with a series of latest and representative dataset distillation methods including DC~\cite{zhao2020dataset}, DM~\cite{zhao2023dataset}, DSA~\cite{zhao2021dataset}, CAFE~\cite{wang2022cafe}, MTT~\cite{cazenavette2022dataset}, 
SMDD~\cite{li2022dataset}
, HaBa~\cite{liu2022dataset}, FTD~\cite{du2023minimizing}, DATM~\cite{guo2023towards} and SeqMatch~\cite{du2023sequential}.
Among them, SMDD is the only dataset distillation method for medical, which follows the work of MTT and uses parameter pruning to remove difficult to match parameters to ensure the robustness of synthesized dataset.
We align with prior works in order to ensure a fair evaluation, we
employ networks with instance normalization by default, and a 3-layer ConvNet~\cite{liu2022convnet} is used as the default distillation architecture, which can maximally adapt to the diversity of medical datasets and~\cite{smith2023convnets} demonstrates its superior performance.
We also evaluate the cross-architecture generalization performance of distilled images on other three standard deep network architectures: AlexNet~\cite{alom2018history}, LeNet~\cite{lecun2015lenet} and ResNet18~\cite{he2016deep}.

\subsection{Implementation Details}

To ensure fairness in comparison, we follow up the experimental setup as stated in~\cite{cui2022dc,lei2023comprehensive}.
In the buffer stage, we generate flat expert trajectories using the method similar to FTD~\cite{du2023minimizing}, and we also use the same suite of differentiable augmentations similar to DSA~\cite{zhao2021dataset} in the distillation and evaluation stage. 
Finally, in the evaluation stage after the synthetic dataset is generated, we train $10$ randomly initialized network on it each time for $1000$ training epochs using SGD optimizer, and report the mean and standard deviation of their accuracy on the real test set.
Notice that all experiments were run on the server with four NVIDIA A40 (48G) GPUs.

\begin{table*}
\centering
\caption{Performances of ConvNet using various distillation methods on the \textbf{High-Resolution} medical datasets.}
\label{table_2}
\renewcommand{\arraystretch}{1.2} 
\resizebox{0.95\linewidth}{!}{
\begin{tabular}{c|cc|cc|cc}
\hline & \multicolumn{2}{|c}{ COVID19-CXR $(112\times112)$ } & \multicolumn{2}{|c}{ BREAST-ULS $(112\times112)$ } & \multicolumn{2}{|c}{ SKIN-HAM $(112\times112)$ } \\
IPC & 2  & {10} & 2 & {10} & 2 & {10} \\ 
\hline Real Dataset & \multicolumn{2}{|c}{$90.22 \pm 0.01$} &  \multicolumn{2}{|c}{$74.00 \pm 0.07 $} & \multicolumn{2}{|c}{$70.17 \pm 0.02 $} \\ 
\hline 
DC~\cite{zhao2020dataset} & $50.38 \pm 0.04$ & $55.29 \pm 0.01$ & $52.40 \pm 0.08$ & $55.00 \pm 0.03$ & $22.01 \pm 0.01$ & $36.84 \pm 0.03$\\
DM~\cite{zhao2023dataset} & $49.84 \pm 0.03$ & $54.01 \pm 0.01$ & $50.07 \pm 0.03$ & $54.93 \pm 0.01$ & $34.74 \pm 0.01$ & $49.90 \pm 0.03$\\
DSA~\cite{zhao2021dataset} & $57.51 \pm 0.01$ & $58.71 \pm 0.01$ & $51.93 \pm 0.07$ & $55.13 \pm 0.02$ & $38.84 \pm 0.03$ & $44.10 \pm 0.02$\\
CAFE~\cite{wang2022cafe} & $53.20 \pm 0.03$ & $55.78 \pm 0.01$ & $52.57 \pm 0.04$ & $53.80 \pm 0.02$ & $47.16 \pm 0.02$ & $50.35 \pm 0.01$\\
MTT~\cite{cazenavette2022dataset} & $50.27 \pm 0.03$ & $58.68 \pm 0.01$  & $57.67 \pm 0.05$ & $67.27 \pm 0.01$ & $31.33 \pm 0.04$  & $51.81 \pm 0.02$\\
SMDD~\cite{li2022dataset}  & $51.91 \pm 0.03$ & $58.55 \pm 0.01$ & $58.43 \pm 0.06$ & $67.60 \pm 0.01$ & $32.07 \pm 0.04$ & $50.76 \pm 0.02$ \\
HaBa~\cite{liu2022dataset} & $48.01 \pm 0.04$ & $58.73 \pm 0.01$ & $54.60 \pm 0.05$ & $63.53 \pm 0.02$ & $34.28 \pm 0.03$ & $51.50 \pm 0.02$ \\
FTD~\cite{du2023minimizing} & $52.52 \pm 0.02$ & $59.38 \pm 0.01$  & $60.13 \pm 0.04$ & $67.73 \pm 0.01$ & $33.92 \pm 0.04$ & $52.49 \pm 0.01$ \\
DATM~\cite{guo2023towards}  & $54.84 \pm 0.03$ & $61.14 \pm 0.04$ & $60.87 \pm 0.05$ & $68.07 \pm 0.01$ & $33.97 \pm 0.04$ & $52.93 \pm 0.02$ \\
SeqMatch~\cite{du2023sequential} & $48.72 \pm 0.03$ & $61.39 \pm 0.04$ & $55.67 \pm 0.04$ & $65.13 \pm 0.02$ & $24.88 \pm 0.04$ & $51.81 \pm 0.01$ \\
\hline 
OUR & $\mathbf{66.18} \pm \mathbf{0.02}$ & $\mathbf{69.65} \pm \mathbf{0.01}$ & $\mathbf{65.37} \pm \mathbf{0.02}$ & $\mathbf{68.90} \pm \mathbf{0.01}$ & $\mathbf{51.19} \pm \mathbf{0.02}$ & $\mathbf{53.85} \pm \mathbf{0.01}$ \\
\hline
\end{tabular}
}
\end{table*}

\subsection{Results}

For the evaluation of distillation performance, we
first generate synthetic datasets through candidate methods and train target networks using these datasets. Then, the performance of trained models is evaluated by the corresponding test set of the original dataset. 

The distillation performance results for high-resolution and mnist 2D/3D medical datasets are shown in Table~\ref{table_2} and Table~\ref{table_6}, respectively.
It is seen that our proposed method outperforms all state-of-the-art baseline methods in each medical dataset with a significant improvement. \textit{e.g.,} we significantly improve the performance on COVID19-CXR by more than $13.9\%$ on average. Similarly, on PathMNIST, the performance improvement is more substantial, exceeding $12.6\%$ on average.
Particularly, compared to the SMDD using the same chest X-ray dataset, we achieved an average improvement of $12.7\%$.
In addition, we also compared our method with $ipc=2$ to the baseline methods with $ipc=2, 4, 6$ on PathMNIST dataset, as shown in Table~\ref{table_3}. It can be intuitively seen that the performance of our synthetic dataset with $ipc=2$ is similar to that of the other baseline methods with $ipc=6$, far exceeding the performance of the same $ipc=2$.

\begin{figure}[htbp]
        \centering
        \subcaptionbox{Time complexity\label{fig_1:a}}{
        \includegraphics[width=1.65in]{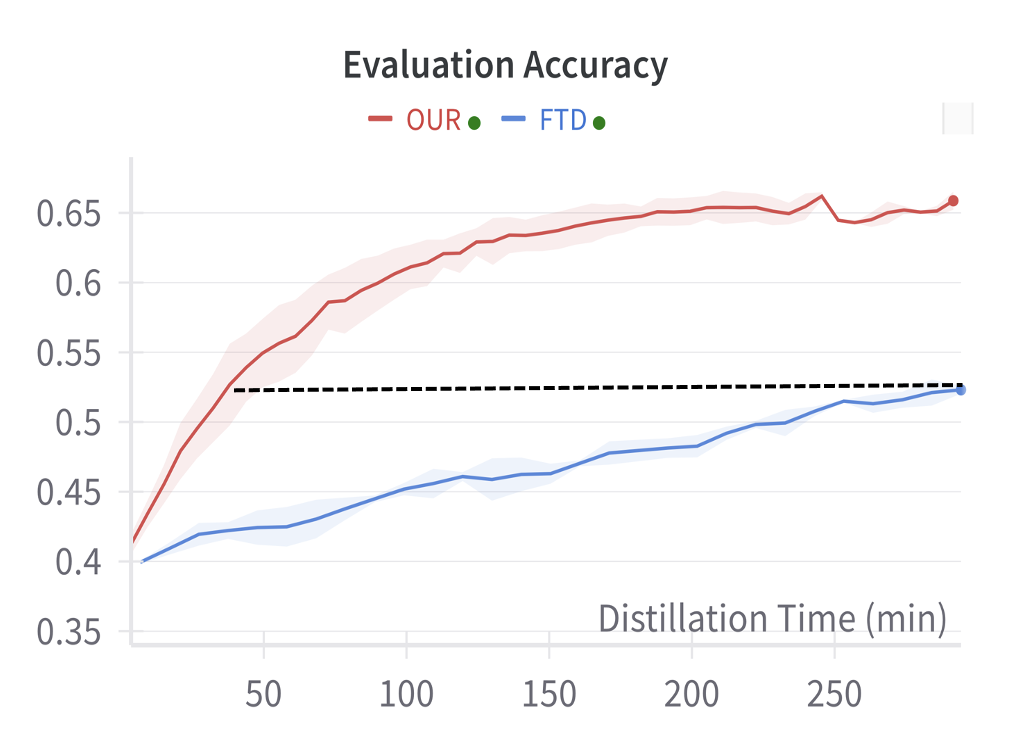}
    }\hspace{-0.5cm}
        \subcaptionbox{Memory complexity\label{fig_1:b}}{
        \centering
        \includegraphics[width=1.75in]{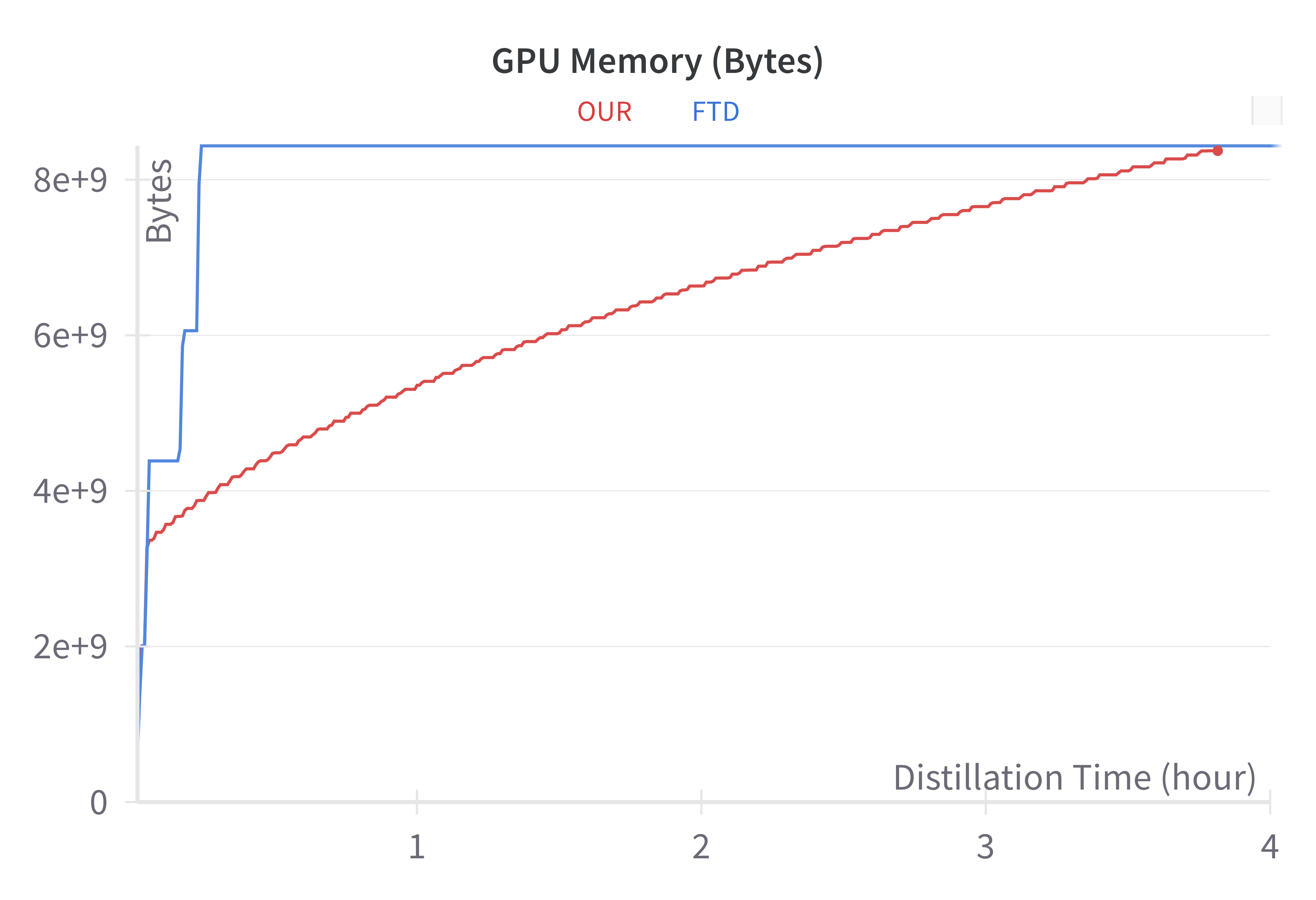}
    }
  \caption{Comparison of the time and memory complexity of our method with a dynamic expert time-step range to other trajectory matching method in the PATHMNIST.}
  \label{fig_1}
\end{figure}

From the performance results, we have summarized the following observations: (1) the dataset distillation methods based on trajectory matching are generally superior to the methods based on distribution matching, especially when the ipc is low.
(2) due to the diversity of medical datasets, multi-step trajectory matching methods are not always better than single-step matching methods. \textit{e.g.,} Seqmatch~\cite{du2023sequential} is not suitable for situations with low ipc, and HaBa~\cite{liu2022dataset} stability is relatively poor.
(3) When the resolution is high, our method can achieve extremely high performance with a low ipc, and this is because the high-resolution synthetic images allow our method to attach more diverse information on them. 
(4) our method has limited improvement when the ipc is high, and this is because a certain number of synthetic images are initialized from the original dataset, which inherently possess considerable diversity and the ability to alleviate overlap.
\textit{i.e.,} the additional information value gained by our method is much lower than the information value added when ipc is low.
\begin{table}
\centering
\small 
\caption{Generalization performances of synthetic datasets on the unseen evaluation models in the PATHMNIST.}
\label{table_5}
\renewcommand{\arraystretch}{1.2} 
\resizebox{1.0\linewidth}{!}{
\begin{tabular}{c|cccc}
\hline & \multicolumn{4}{|c}{ Evaluation Models } \\
Method & ConvNet & LeNet & ResNet18 & AlexNet \\
\hline
DC & $57.63 \pm 0.01$ & $44.35 \pm 0.03$ & $34.22 \pm 0.02$ & $40.72 \pm 0.05$ \\
DM & $50.74 \pm 0.02$ & $41.59 \pm 0.04$ & $42.84 \pm 0.03$ & $41.25 \pm 0.04$ \\
DSA & $58.56 \pm 0.02$ & $51.26 \pm 0.05$ & $43.63 \pm 0.02$ & $53.85 \pm 0.04$ \\
MTT & $50.86 \pm 0.02$ & $44.95 \pm 0.03$ & $44.71 \pm 0.04$ & $45.11 \pm 0.05$ \\
FTD & $53.47 \pm 0.02$ & $44.21 \pm 0.02$ & $45.87 \pm 0.04$ & $46.27 \pm 0.02$ \\
OUR & $\mathbf{67.01} \pm \mathbf{0.01}$ & $\mathbf{51.64} \pm \mathbf{0.03}$ & $\mathbf{52.46} \pm \mathbf{0.03}$ & $\mathbf{53.72} \pm \mathbf{0.03}$ \\
\hline
\end{tabular}
}
\end{table}


 \section{Analysis}

\subsection{Quantitative Analysis}

\subsubsection{Cross-Architecture Generalization.}
For dataset distillation methods that need to be deployed in practical application scenarios, a satisfactory distilled dataset should have \textit{similar training effects to the original one on downstream models with arbitrary architectures}. 
Therefore, cross-architecture generalization performance is an important metric, and we report the generalization performances of our method and competitors in the PATHMNIST on Table~\ref{table_5}. 
The results indicate that compared to existing dataset distillation methods, our proposed method still maintains the best performance when faced with network architectures different from those used in buffer and distillation stages.

\begin{figure}[htbp]
        \centering
        \subcaptionbox{Trajectory matching loss\label{fig:2:a}}{
        \includegraphics[width=1.57in]{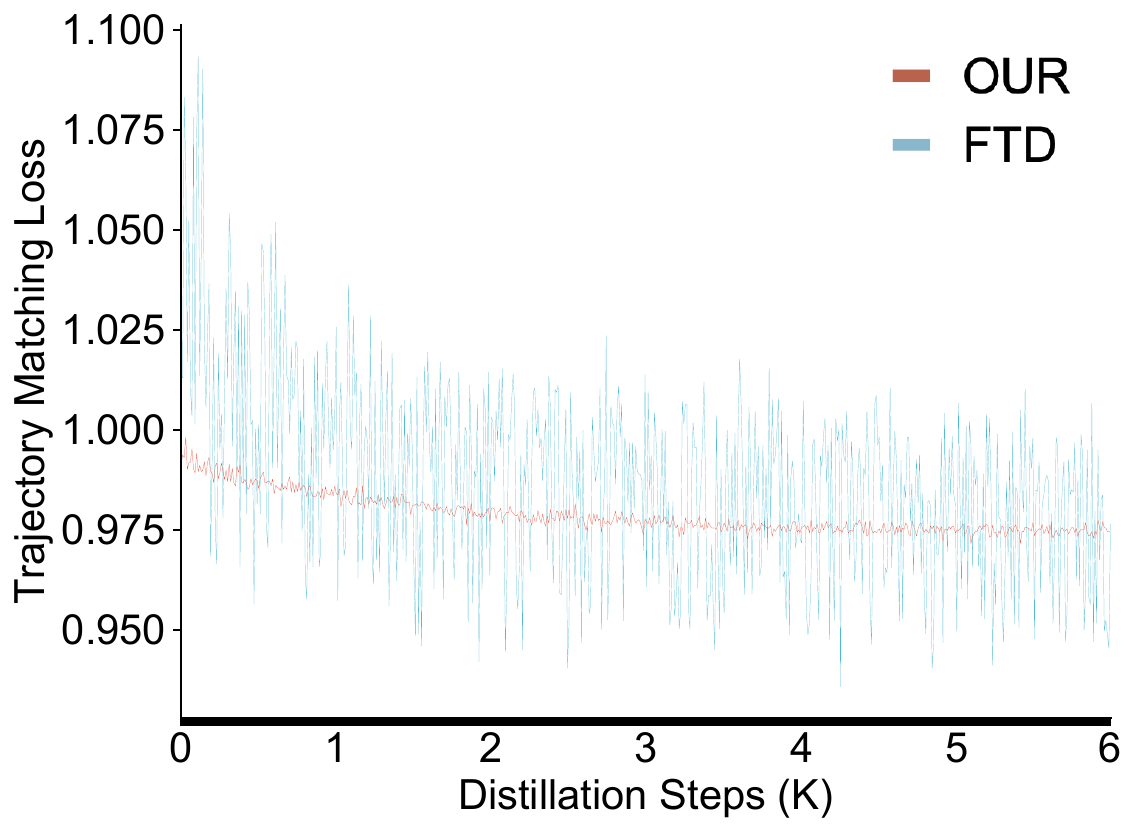}
    }\hspace{-0.25cm}
        \subcaptionbox{Evaluation stability variance\label{fig:2:b}}{
        \centering
        \includegraphics[width=1.58in]{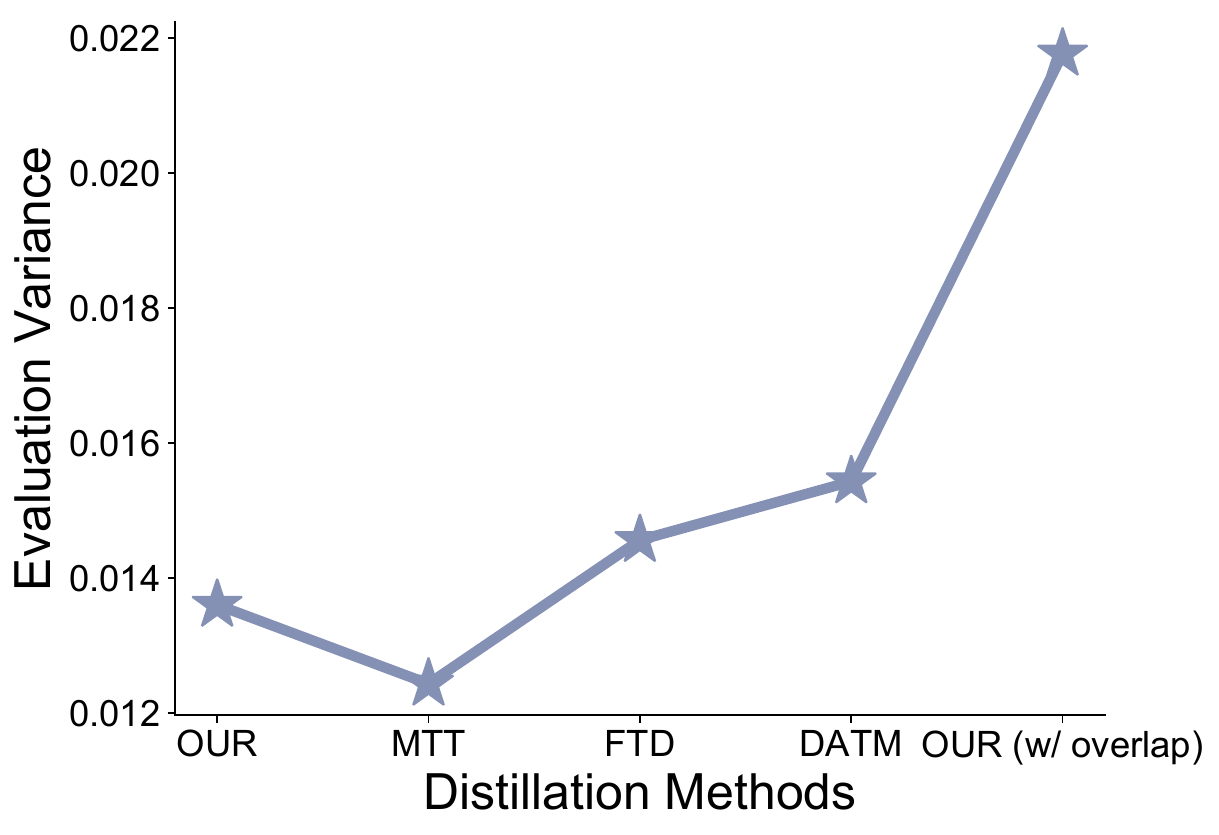}
    }
 \caption{Comparison of the stability in the PATHMNIST during distillation and evaluation stages.
 }
 \label{fig:2}
\end{figure}

\subsubsection{Time and Memory Complexity Evaluation}
We evaluate the running time and GPU memory of mainstream trajectory matching methods to estimate the training cost brought by distillation.
Specifically, we first analyze the evaluation accuracy of our and FTD methods over the distillation duration, and the expert time-step range of the FTD is set to 50.
Since our method uses a dynamic expert time-step range that gradually increases from $1$, we can achieve performance comparable to FTD in very little time. 
As shown in Figure~\ref{fig_1:a}, we achieved $100\%$ performance of FTD using approximately one-sixth of the duration. Similarly, we also analyzed the GPU memory usage of our method and FTD in Figure~\ref{fig_1:b}, and our method has consistently consumed less graphics memory than FTD, especially in the early stages of training.
Note that we evaluate these two methods in the same computation environment and fairly.


\subsection{Ablation Study}
In this section, we conduct a series of ablation studies for our method to investigate the effectiveness of the proposed two main contributions: progressive stable distillation, dynamic retraining and overlap mitigation.

\textbf{Progressive Stable Distillation.}
To verify the stability of our method, we evaluated the trajectory-matching loss in the distillation stage and the stability variance in the evaluation stage. As shown in Figure~\ref{fig:2:a}, it can be observed that our method achieves more stable loss compared to FTD due to the progressive training strategy.
As shown in Figure~\ref{fig:2:b}, we evaluate the stability of the synthetic datasets distilled by different trajectory matching methods using $10$ randomly initialized models. It can be observed that the synthetic dataset generated by our method has good evaluation stability after retraining and overlap mitigation.

\begin{figure}[htbp]
    \centering
    \subcaptionbox{MMD\label{fig:3:c}}{
        \includegraphics[width=1.105in]{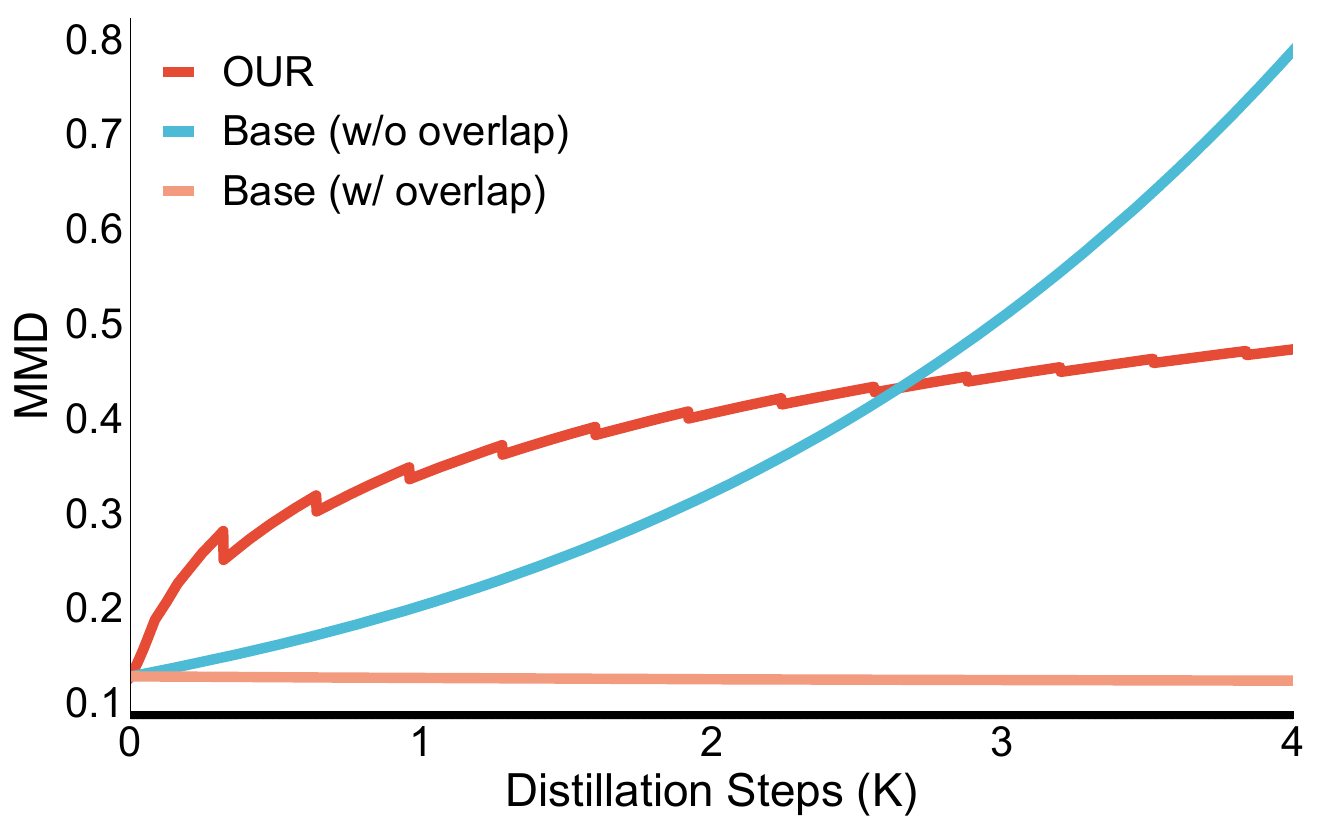}  
    }\hspace{-0.2cm}
        \subcaptionbox{IPC=2\label{fig:3:a}}{
        \includegraphics[width=1.107in]{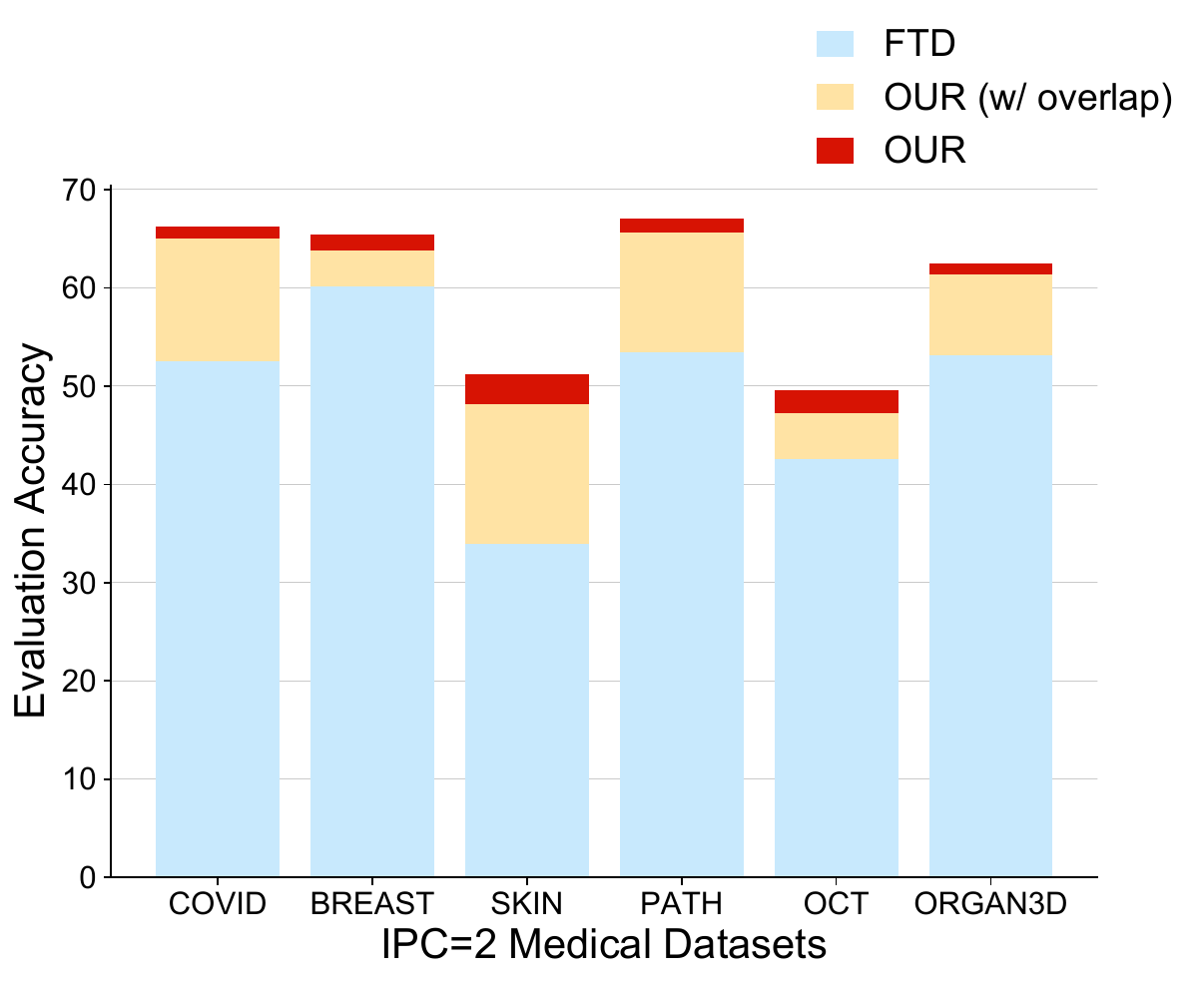}
    }\hspace{-0.35cm}
        \subcaptionbox{IPC=10\label{fig:3:b}}{
        \includegraphics[width=1.107in]{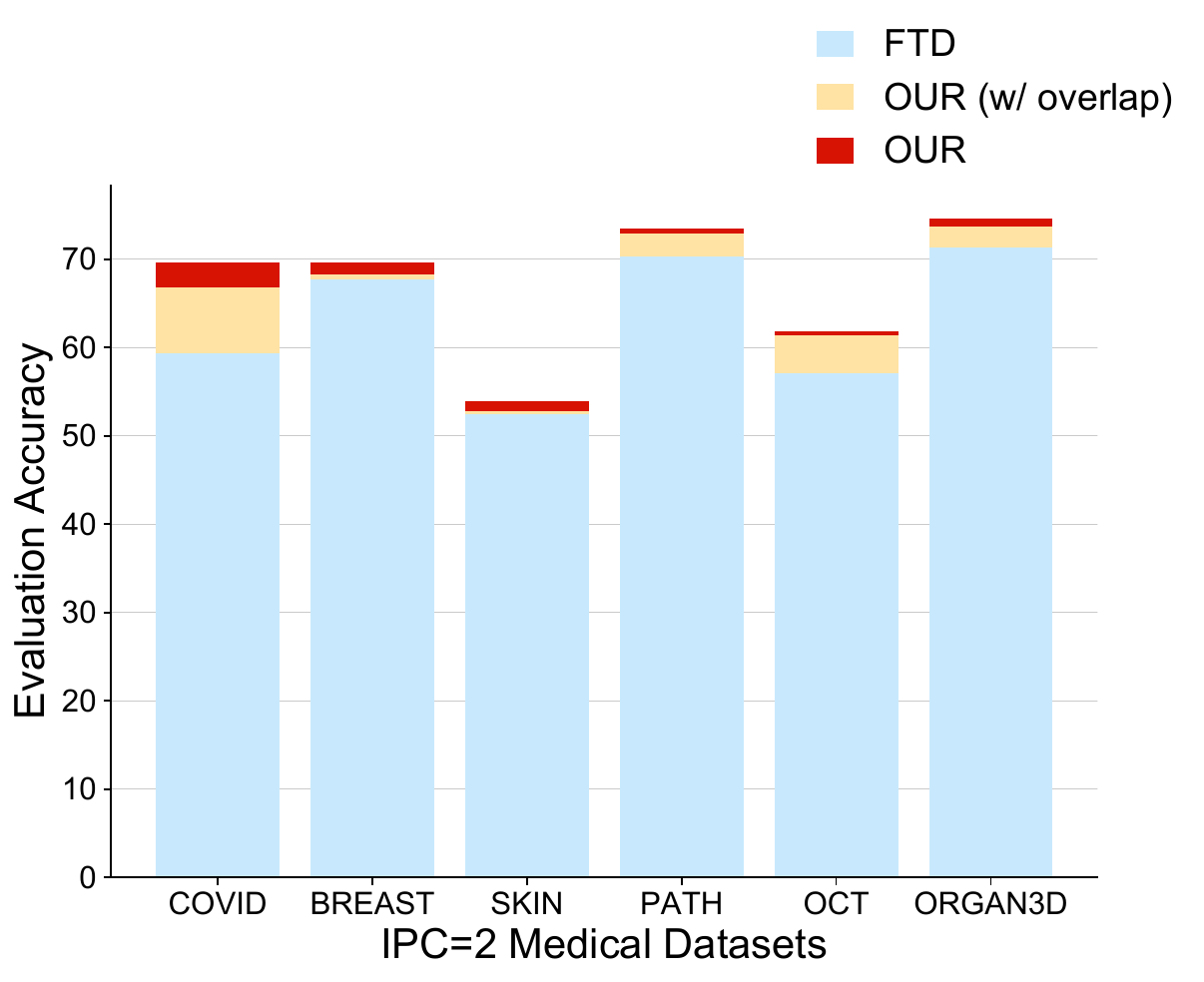}
    }      
    \caption{(a) MMD plots using different overlap mitigation losses in the PATHMNIST. (b) (c) visualization of the performance improvements in the ablation study.}
        \label{fig:3}
\end{figure}
\begin{table}
\centering
\small 
\caption{Comparison of the superior performances of our method with a fixed fewer IPC to other dataset distillation methods in the 
PATHMNIST.}
\label{table_3}
\renewcommand{\arraystretch}{1.2}
\resizebox{1.0\linewidth}{!}{
\begin{tabular}{c|cccc}
\hline & \multicolumn{4}{|c}{ Evaluation Methods } \\
IPC & DC & DM & MTT & FTD \\
\hline
2 & $57.63 \pm 0.01 $ & $50.74 \pm 0.02 $ & $ 50.86 \pm 0.02 $ & $53.47 \pm 0.02$ \\
4 & $63.23 \pm 0.01 $ & $60.42 \pm 0.02 $ & $61.97 \pm 0.02 $ & $66.33 \pm 0.01 $ \\
6 & $67.61 \pm 0.01 $ & $68.42 \pm 0.01 $ & $66.54 \pm 0.04 $ & $68.38 \pm 0.01 $ \\
\hline
OUR$_{IPC=2}$ & \multicolumn{4}{|c}{ $\mathbf{67.01} \pm \mathbf{0.01}$ } \\
\hline
\end{tabular}
}
\end{table}

\textbf{Dynamic Retraining and Overlap Mitigation.} We use MMD to demonstrate the existence of overlap in the distillation stage and the necessity of overlap mitigation.
The orange line in Figure~\ref{fig:3:c} shows that as the distillation progresses, the MMD among images of the same class continues to decrease, indicating that they become increasingly similar.
The blue line indicates that the MMD continues to increase and does not converge when we only add the overlap mitigation loss, as this loss will dominate at the end of distillation.
The red line shows that after adding the dynamic retraining components, the MMD can maintain increased and gradually converged. It increases the diversity of synthetic images while maintaining the stability.
In addition, the improvements of different components in our method are in Figure~\ref{fig:3} (b)(c), showing their effectiveness.

\begin{figure}[htbp]
        \centering
        \subcaptionbox{Buffer trajectory}{
        \includegraphics[width=1.11in]{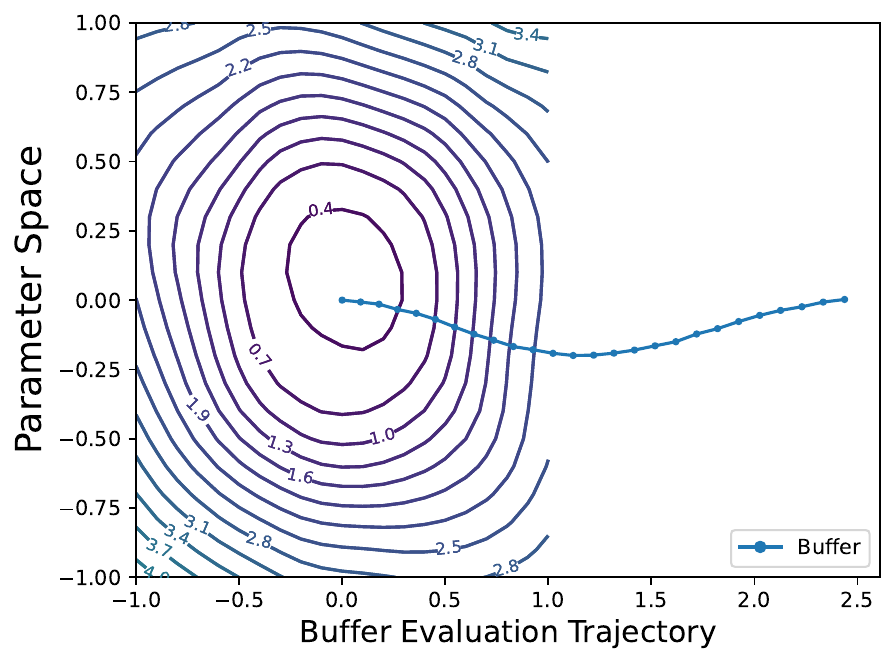}
    }\hspace{-0.3cm}
        \subcaptionbox{FTD trajectory}{
        \centering
        \includegraphics[width=1.11in]{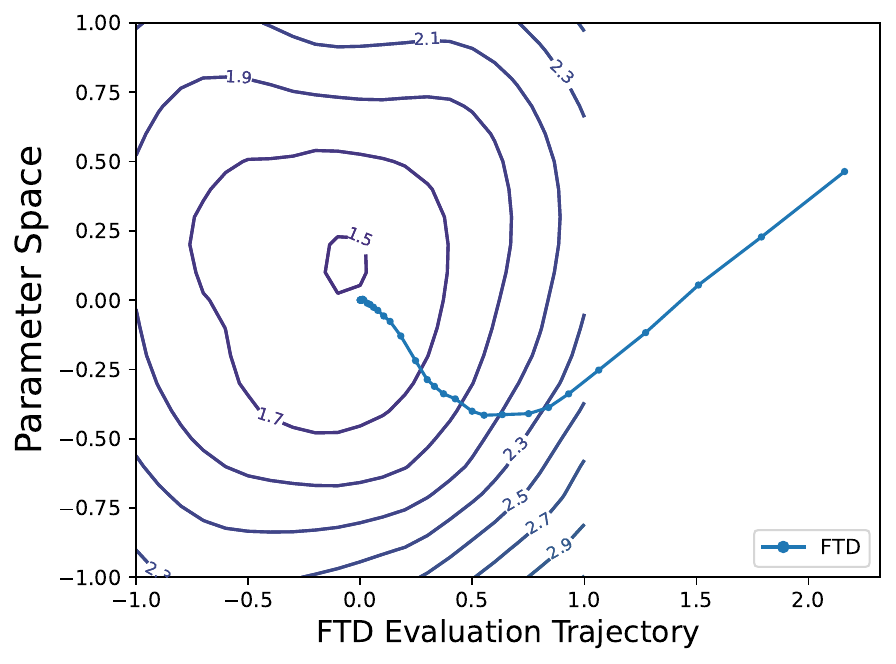}
    }\hspace{-0.3cm}
    \subcaptionbox{OUR trajectory}{
        \centering
        \includegraphics[width=1.11in]{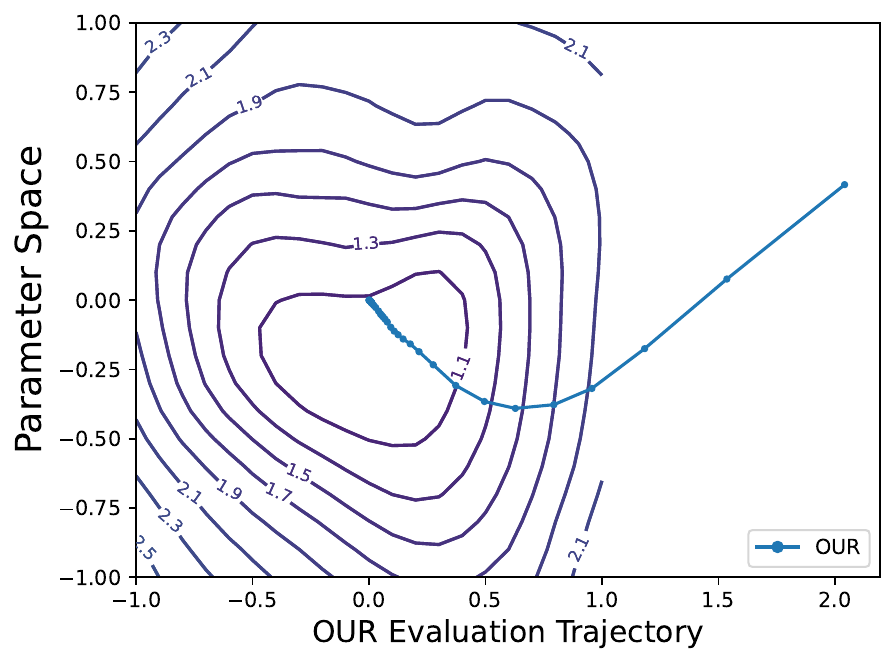}
    }
    \subcaptionbox{Buffer surface}{
    \centering
        \includegraphics[width=1.05in]{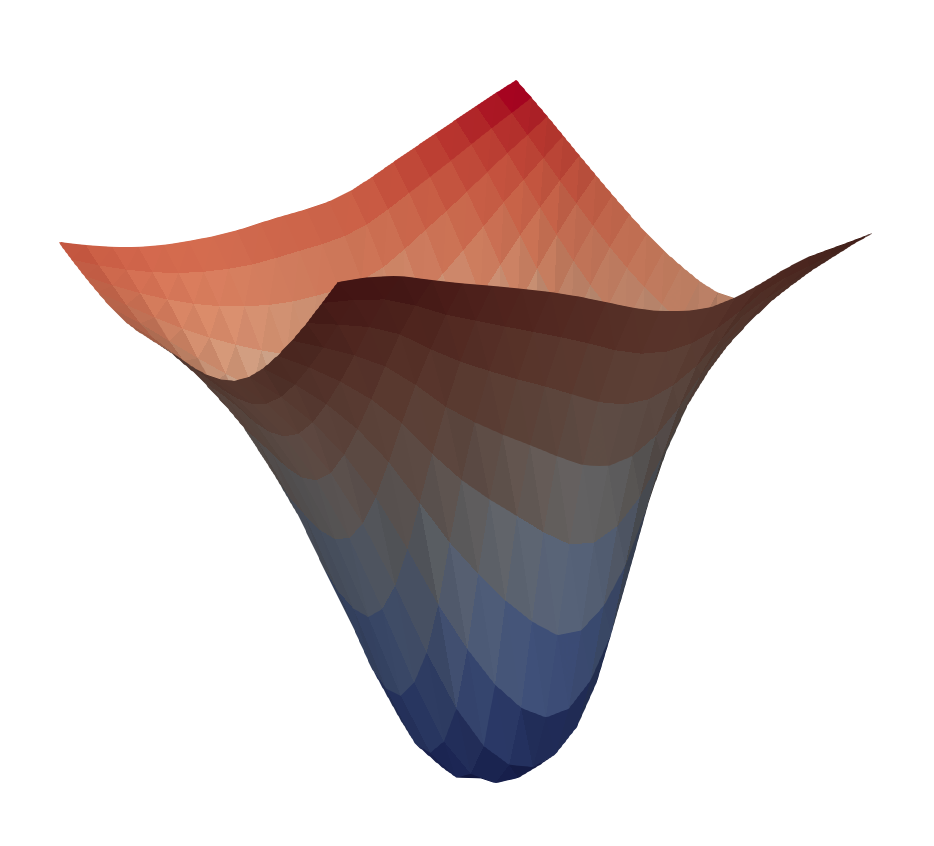}
    }\hspace{-0.1cm}
        \subcaptionbox{FTD surface}{
        \centering
        \includegraphics[width=1.05in]{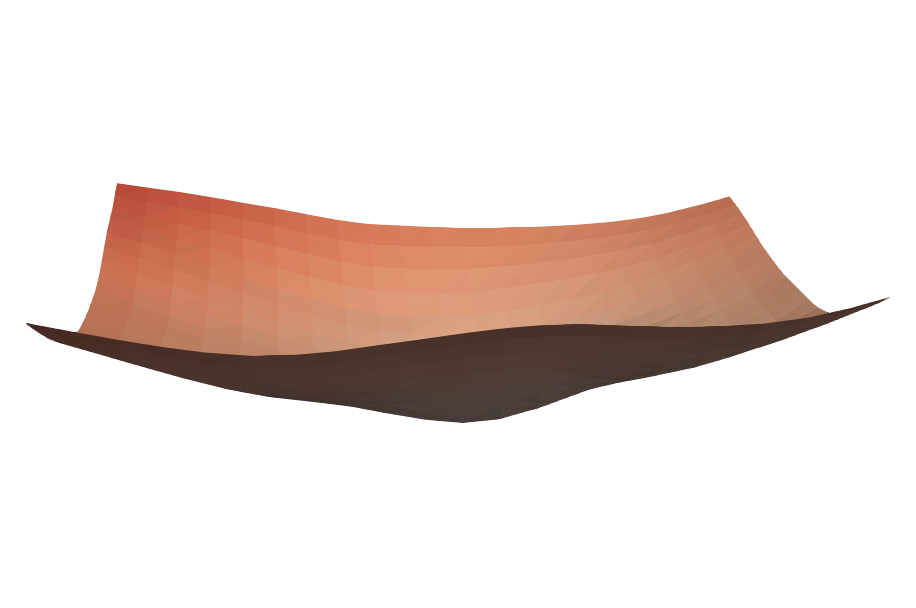}
    }\hspace{-0.1cm}
    \subcaptionbox{OUR surface}{
        \centering
        \includegraphics[width=1.07in]{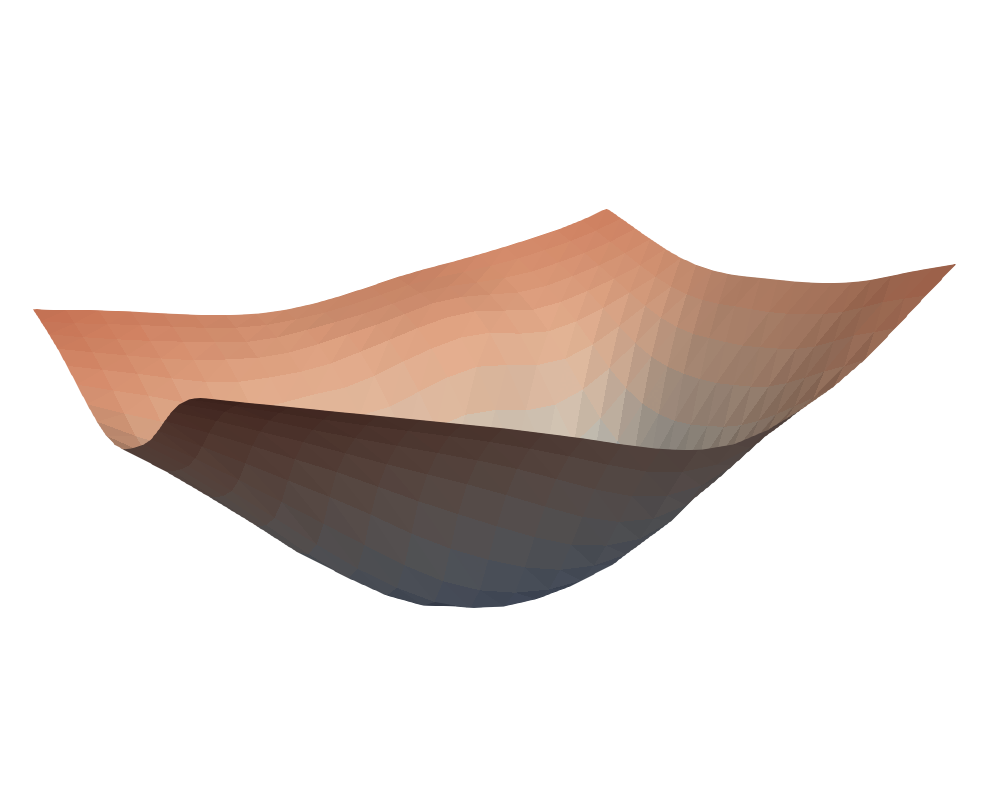}
    }
        \caption{Visualization of 2D/3D evaluation trajectory using loss landscape in the PATHMNIST.}
        \label{fig:10}
\end{figure}

\begin{figure}[htbp]
        \centering
        \subcaptionbox{Buffer}{
        \includegraphics[width=1.181in]{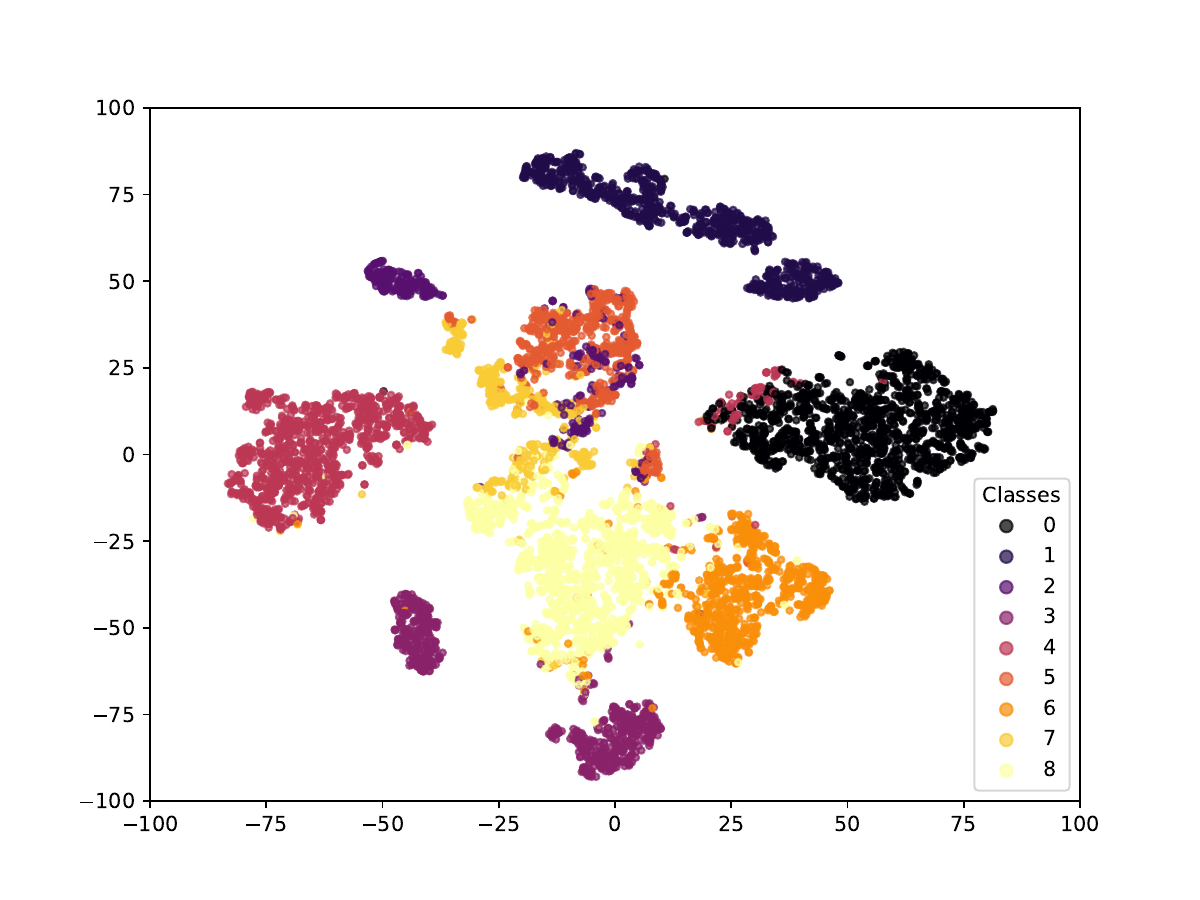}
    }\hspace{-0.56cm}
        \subcaptionbox{FTD}{
        \centering
        \includegraphics[width=1.181in]{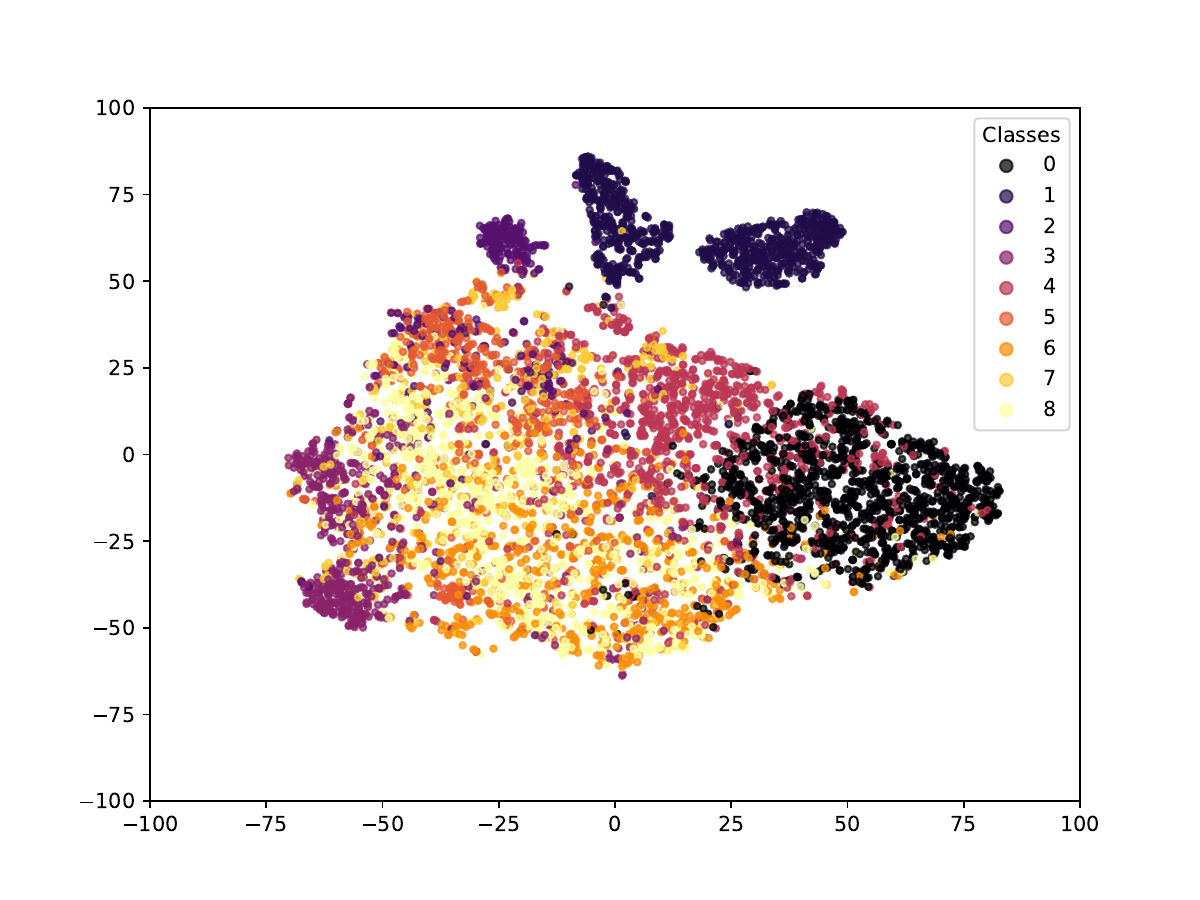}
    }\hspace{-0.56cm}
        \subcaptionbox{OUR}{
        \includegraphics[width=1.181in]{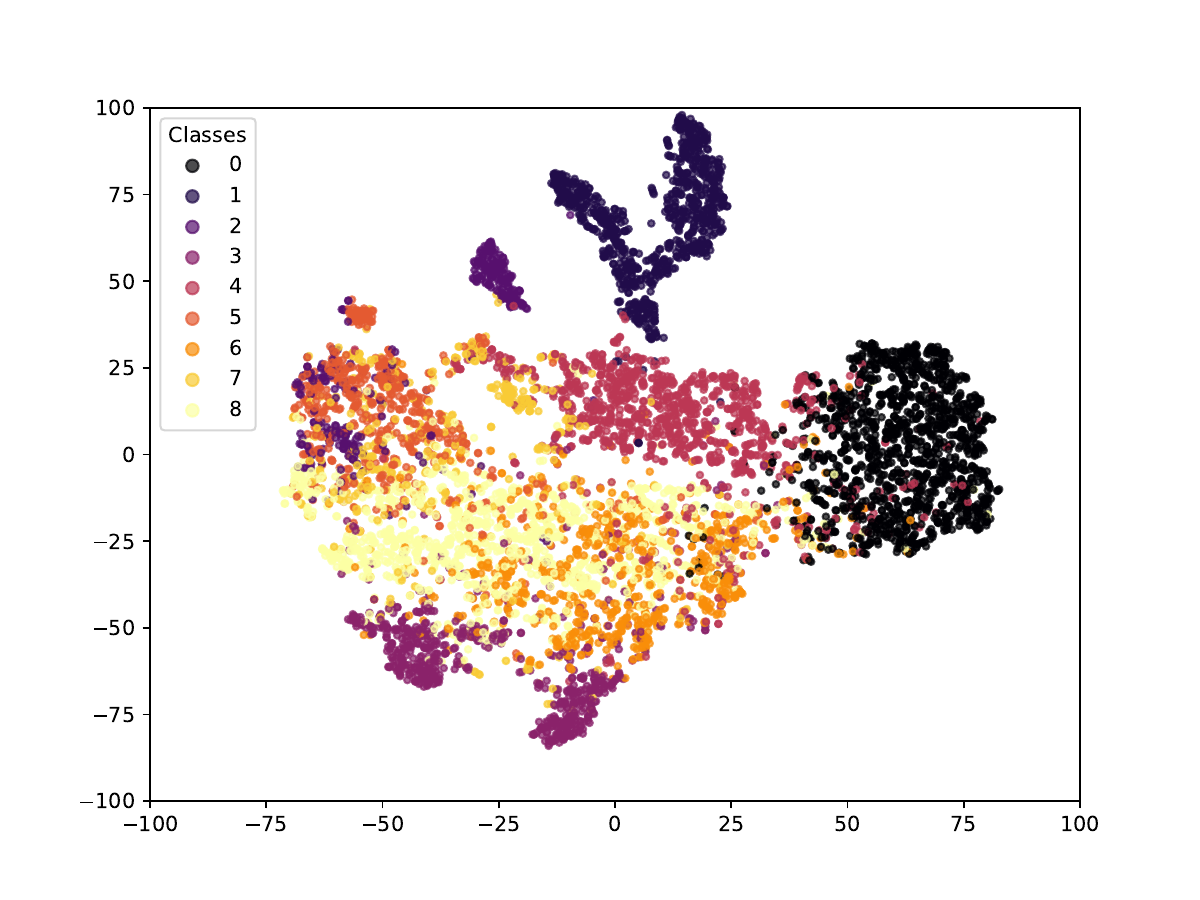}
    }
    \caption{Feature visualization using t-SNE in the PATHMNIST during buffer and evaluation stages.}
    \label{fig:11}
\end{figure}

\subsection{Qualitative Analysis}

In this section, we perform loss landscape~\cite{li2018visualizing} and t-SNE~\cite{11van2008visualizing} respectively to visualize the performance of trajectory matching and features.
As shown in Figure~\ref{fig:10}, we present the evaluation trajectories of different methods on dynamic 2D contour loss curves and static 3D loss landscapes respectively. We can observe that our method has denser contour lines, a higher and narrower landscape compared to FTD, which verifies that our trajectory converges and matches better to the buffer~\cite{garipov2018loss,garipov2018loss}.
As shown in Figure~\ref{fig:11}, we demonstrate that our method achieved fewer inter-class confusion issues than FTD. More analysis results are in the supplementary materials.


 \section{Conclusions}
This paper established a new and comprehensive medical image dataset distillation benchmark. Through evaluation, the proposed progressive trajectory matching strategy and the overlap elimination of synthetic images achieve SOTA performances.  

\bibliographystyle{named}
\bibliography{ijcai24}

\begin{thebibliography}{}

\bibitem[\protect\citeauthoryear{Al-Dhabyani \bgroup \em et al.\egroup }{2020}]{al2020dataset}
Walid Al-Dhabyani, Mohammed Gomaa, Hussien Khaled, and Aly Fahmy.
\newblock Dataset of breast ultrasound images.
\newblock {\em Data in brief}, 28:104863, 2020.

\bibitem[\protect\citeauthoryear{Alom \bgroup \em et al.\egroup }{2018}]{alom2018history}
Md~Zahangir Alom, Tarek~M Taha, Christopher Yakopcic, Stefan Westberg, Paheding Sidike, Mst~Shamima Nasrin, Brian~C Van~Esesn, Abdul A~S Awwal, and Vijayan~K Asari.
\newblock The history began from alexnet: A comprehensive survey on deep learning approaches.
\newblock {\em arXiv preprint arXiv:1803.01164}, 2018.

\bibitem[\protect\citeauthoryear{Bilic \bgroup \em et al.\egroup }{2023}]{bilic2023liver}
Patrick Bilic, Patrick Christ, Hongwei~Bran Li, Eugene Vorontsov, Avi Ben-Cohen, Georgios Kaissis, Adi Szeskin, Colin Jacobs, Gabriel Efrain~Humpire Mamani, Gabriel Chartrand, et~al.
\newblock The liver tumor segmentation benchmark (lits).
\newblock {\em Medical Image Analysis}, 84:102680, 2023.

\bibitem[\protect\citeauthoryear{Cazenavette \bgroup \em et al.\egroup }{2022}]{cazenavette2022dataset}
George Cazenavette, Tongzhou Wang, Antonio Torralba, Alexei~A Efros, and Jun-Yan Zhu.
\newblock Dataset distillation by matching training trajectories.
\newblock In {\em Proceedings of the IEEE/CVF Conference on Computer Vision and Pattern Recognition}, pages 4750--4759, 2022.

\bibitem[\protect\citeauthoryear{Cui \bgroup \em et al.\egroup }{2022}]{cui2022dc}
Justin Cui, Ruochen Wang, Si~Si, and Cho-Jui Hsieh.
\newblock Dc-bench: Dataset condensation benchmark.
\newblock {\em Advances in Neural Information Processing Systems}, 35:810--822, 2022.

\bibitem[\protect\citeauthoryear{Du \bgroup \em et al.\egroup }{2023a}]{du2023minimizing}
Jiawei Du, Yidi Jiang, Vincent~YF Tan, Joey~Tianyi Zhou, and Haizhou Li.
\newblock Minimizing the accumulated trajectory error to improve dataset distillation.
\newblock In {\em Proceedings of the IEEE/CVF Conference on Computer Vision and Pattern Recognition}, pages 3749--3758, 2023.

\bibitem[\protect\citeauthoryear{Du \bgroup \em et al.\egroup }{2023b}]{du2023sequential}
Jiawei Du, Qin Shi, and Joey~Tianyi Zhou.
\newblock Sequential subset matching for dataset distillation.
\newblock {\em arXiv preprint arXiv:2311.01570}, 2023.

\bibitem[\protect\citeauthoryear{Garipov \bgroup \em et al.\egroup }{2018}]{garipov2018loss}
Timur Garipov, Pavel Izmailov, Dmitrii Podoprikhin, Dmitry~P Vetrov, and Andrew~G Wilson.
\newblock Loss surfaces, mode connectivity, and fast ensembling of dnns.
\newblock {\em Advances in neural information processing systems}, 31, 2018.

\bibitem[\protect\citeauthoryear{Geng \bgroup \em et al.\egroup }{2023}]{geng2023survey}
Jiahui Geng, Zongxiong Chen, Yuandou Wang, Herbert Woisetschlaeger, Sonja Schimmler, Ruben Mayer, Zhiming Zhao, and Chunming Rong.
\newblock A survey on dataset distillation: Approaches, applications and future directions.
\newblock {\em arXiv preprint arXiv:2305.01975}, 2023.

\bibitem[\protect\citeauthoryear{Gretton \bgroup \em et al.\egroup }{2012}]{gretton2012kernel}
Arthur Gretton, Karsten~M Borgwardt, Malte~J Rasch, Bernhard Sch{\"o}lkopf, and Alexander Smola.
\newblock A kernel two-sample test.
\newblock {\em The Journal of Machine Learning Research}, 13(1):723--773, 2012.

\bibitem[\protect\citeauthoryear{Guo \bgroup \em et al.\egroup }{2023}]{guo2023towards}
Ziyao Guo, Kai Wang, George Cazenavette, Hui Li, Kaipeng Zhang, and Yang You.
\newblock Towards lossless dataset distillation via difficulty-aligned trajectory matching.
\newblock {\em arXiv preprint arXiv:2310.05773}, 2023.

\bibitem[\protect\citeauthoryear{He \bgroup \em et al.\egroup }{2016}]{he2016deep}
Kaiming He, Xiangyu Zhang, Shaoqing Ren, and Jian Sun.
\newblock Deep residual learning for image recognition.
\newblock In {\em Proceedings of the IEEE conference on computer vision and pattern recognition}, pages 770--778, 2016.

\bibitem[\protect\citeauthoryear{Kather \bgroup \em et al.\egroup }{2019}]{kather2019predicting}
Jakob~Nikolas Kather, Johannes Krisam, Pornpimol Charoentong, Tom Luedde, Esther Herpel, Cleo-Aron Weis, Timo Gaiser, Alexander Marx, Nektarios~A Valous, Dyke Ferber, et~al.
\newblock Predicting survival from colorectal cancer histology slides using deep learning: A retrospective multicenter study.
\newblock {\em PLoS medicine}, 16(1):e1002730, 2019.

\bibitem[\protect\citeauthoryear{Kermany \bgroup \em et al.\egroup }{2018}]{kermany2018identifying}
Daniel~S Kermany, Michael Goldbaum, Wenjia Cai, Carolina~CS Valentim, Huiying Liang, Sally~L Baxter, Alex McKeown, Ge~Yang, Xiaokang Wu, Fangbing Yan, et~al.
\newblock Identifying medical diagnoses and treatable diseases by image-based deep learning.
\newblock {\em cell}, 172(5):1122--1131, 2018.

\bibitem[\protect\citeauthoryear{LeCun and others}{2015}]{lecun2015lenet}
Yann LeCun et~al.
\newblock Lenet-5, convolutional neural networks.
\newblock {\em URL: http://yann. lecun. com/exdb/lenet}, 20(5):14, 2015.

\bibitem[\protect\citeauthoryear{Lei and Tao}{2023}]{lei2023comprehensive}
Shiye Lei and Dacheng Tao.
\newblock A comprehensive survey to dataset distillation.
\newblock {\em arXiv preprint arXiv:2301.05603}, 2023.

\bibitem[\protect\citeauthoryear{Li \bgroup \em et al.\egroup }{2018}]{li2018visualizing}
Hao Li, Zheng Xu, Gavin Taylor, Christoph Studer, and Tom Goldstein.
\newblock Visualizing the loss landscape of neural nets.
\newblock {\em Advances in neural information processing systems}, 31, 2018.

\bibitem[\protect\citeauthoryear{Li \bgroup \em et al.\egroup }{2020}]{li2020soft}
Guang Li, Ren Togo, Takahiro Ogawa, and Miki Haseyama.
\newblock Soft-label anonymous gastric x-ray image distillation.
\newblock In {\em 2020 IEEE International Conference on Image Processing (ICIP)}, pages 305--309. IEEE, 2020.

\bibitem[\protect\citeauthoryear{Li \bgroup \em et al.\egroup }{2022a}]{li2022compressed}
Guang Li, Ren Togo, Takahiro Ogawa, and Miki Haseyama.
\newblock Compressed gastric image generation based on soft-label dataset distillation for medical data sharing.
\newblock {\em Computer Methods and Programs in Biomedicine}, 227:107189, 2022.

\bibitem[\protect\citeauthoryear{Li \bgroup \em et al.\egroup }{2022b}]{li2022dataset}
Guang Li, Ren Togo, Takahiro Ogawa, and Miki Haseyama.
\newblock Dataset distillation for medical dataset sharing.
\newblock {\em arXiv preprint arXiv:2209.14603}, 2022.

\bibitem[\protect\citeauthoryear{Liu \bgroup \em et al.\egroup }{2022a}]{liu2022dataset}
Songhua Liu, Kai Wang, Xingyi Yang, Jingwen Ye, and Xinchao Wang.
\newblock Dataset distillation via factorization.
\newblock {\em Advances in Neural Information Processing Systems}, 35:1100--1113, 2022.

\bibitem[\protect\citeauthoryear{Liu \bgroup \em et al.\egroup }{2022b}]{liu2022convnet}
Zhuang Liu, Hanzi Mao, Chao-Yuan Wu, Christoph Feichtenhofer, Trevor Darrell, and Saining Xie.
\newblock A convnet for the 2020s.
\newblock In {\em Proceedings of the IEEE/CVF conference on computer vision and pattern recognition}, pages 11976--11986, 2022.

\bibitem[\protect\citeauthoryear{Rahman \bgroup \em et al.\egroup }{2021}]{rahman2021exploring}
Tawsifur Rahman, Amith Khandakar, Yazan Qiblawey, Anas Tahir, Serkan Kiranyaz, Saad Bin~Abul Kashem, Mohammad~Tariqul Islam, Somaya Al~Maadeed, Susu~M Zughaier, Muhammad~Salman Khan, et~al.
\newblock Exploring the effect of image enhancement techniques on covid-19 detection using chest x-ray images.
\newblock {\em Computers in biology and medicine}, 132:104319, 2021.

\bibitem[\protect\citeauthoryear{Sachdeva and McAuley}{2023}]{sachdeva2023data}
Noveen Sachdeva and Julian McAuley.
\newblock Data distillation: A survey.
\newblock {\em arXiv preprint arXiv:2301.04272}, 2023.

\bibitem[\protect\citeauthoryear{Smith \bgroup \em et al.\egroup }{2023}]{smith2023convnets}
Samuel~L Smith, Andrew Brock, Leonard Berrada, and Soham De.
\newblock Convnets match vision transformers at scale.
\newblock {\em arXiv preprint arXiv:2310.16764}, 2023.

\bibitem[\protect\citeauthoryear{Tschandl \bgroup \em et al.\egroup }{2018}]{tschandl2018ham10000}
Philipp Tschandl, Cliff Rosendahl, and Harald Kittler.
\newblock The ham10000 dataset, a large collection of multi-source dermatoscopic images of common pigmented skin lesions.
\newblock {\em Scientific data}, 5(1):1--9, 2018.

\bibitem[\protect\citeauthoryear{Van~der Maaten and Hinton}{2008}]{11van2008visualizing}
Laurens Van~der Maaten and Geoffrey Hinton.
\newblock Visualizing data using t-sne.
\newblock {\em Journal of machine learning research}, 9(11), 2008.

\bibitem[\protect\citeauthoryear{Wang \bgroup \em et al.\egroup }{2018}]{wang2018dataset}
Tongzhou Wang, Jun-Yan Zhu, Antonio Torralba, and Alexei~A Efros.
\newblock Dataset distillation.
\newblock {\em arXiv preprint arXiv:1811.10959}, 2018.

\bibitem[\protect\citeauthoryear{Wang \bgroup \em et al.\egroup }{2022}]{wang2022cafe}
Kai Wang, Bo~Zhao, Xiangyu Peng, Zheng Zhu, Shuo Yang, Shuo Wang, Guan Huang, Hakan Bilen, Xinchao Wang, and Yang You.
\newblock Cafe: Learning to condense dataset by aligning features.
\newblock In {\em Proceedings of the IEEE/CVF Conference on Computer Vision and Pattern Recognition}, pages 12196--12205, 2022.

\bibitem[\protect\citeauthoryear{Yang \bgroup \em et al.\egroup }{2023}]{yang2023medmnist}
Jiancheng Yang, Rui Shi, Donglai Wei, Zequan Liu, Lin Zhao, Bilian Ke, Hanspeter Pfister, and Bingbing Ni.
\newblock Medmnist v2-a large-scale lightweight benchmark for 2d and 3d biomedical image classification.
\newblock {\em Scientific Data}, 10(1):41, 2023.

\bibitem[\protect\citeauthoryear{Yu \bgroup \em et al.\egroup }{2023}]{yu2023dataset}
Ruonan Yu, Songhua Liu, and Xinchao Wang.
\newblock Dataset distillation: A comprehensive review.
\newblock {\em arXiv preprint arXiv:2301.07014}, 2023.

\bibitem[\protect\citeauthoryear{Zhao and Bilen}{2021}]{zhao2021dataset}
Bo~Zhao and Hakan Bilen.
\newblock Dataset condensation with differentiable siamese augmentation.
\newblock In {\em International Conference on Machine Learning}, pages 12674--12685. PMLR, 2021.

\bibitem[\protect\citeauthoryear{Zhao and Bilen}{2023}]{zhao2023dataset}
Bo~Zhao and Hakan Bilen.
\newblock Dataset condensation with distribution matching.
\newblock In {\em Proceedings of the IEEE/CVF Winter Conference on Applications of Computer Vision}, pages 6514--6523, 2023.

\bibitem[\protect\citeauthoryear{Zhao \bgroup \em et al.\egroup }{2020}]{zhao2020dataset}
Bo~Zhao, Konda~Reddy Mopuri, and Hakan Bilen.
\newblock Dataset condensation with gradient matching.
\newblock {\em arXiv preprint arXiv:2006.05929}, 2020.

\end{thebibliography}

\end{document}